\documentclass[lettersize,journal]{IEEEtran}
\usepackage[noadjust]{cite}
\usepackage{amsmath,amsfonts}
\usepackage{algorithmic}
\usepackage{array}
\usepackage[caption=false,font=normalsize,labelfont=sf,textfont=sf]{subfig}
\usepackage{textcomp}
\usepackage{stfloats}
\usepackage{url}
\usepackage{verbatim}
\usepackage{graphicx}
\usepackage{booktabs}
\usepackage{times}
\usepackage{epsfig}
\usepackage{graphicx}
\usepackage{amsmath}
\usepackage{amssymb}
\usepackage[ruled]{algorithm2e}
\usepackage{multirow}
\usepackage{indentfirst}
\usepackage{graphicx}
\usepackage{float}
\usepackage{dirtree}
\usepackage{amsmath,amssymb}
\usepackage{setspace}
\usepackage{mathtools}
\usepackage{float}
\usepackage[normalem]{ulem}
\usepackage{array}
\usepackage{epstopdf}
\usepackage{amsmath}
\usepackage{times}
\usepackage{epsfig}
\usepackage{graphicx}
\usepackage{amsmath}
\usepackage{amssymb}
\usepackage{caption}
\usepackage{color}
\usepackage{subfig}
\usepackage{lipsum}
\usepackage{booktabs}
\usepackage{bm}
\usepackage{multicol}
\usepackage{multirow}
\usepackage{bbding}
\usepackage{array}
\usepackage{diagbox}
\usepackage{makecell}
\usepackage{hyperref}

\usepackage[caption=false,font=normalsize,labelfont=sf,textfont=sf]{subfig}
\usepackage{textcomp}
\usepackage{stfloats}
\usepackage{url}
\usepackage{verbatim}
\usepackage{graphicx}
\def\BibTeX{{\rm B\kern-.05em{\sc i\kern-.025em b}\kern-.08em
    T\kern-.1667em\lower.7ex\hbox{E}\kern-.125emX}}
\usepackage{balance}

\begin{document}
\title{MARVEL: Raster Manga Vectorization via Primitive-wise Deep Reinforcement Learning}
\author{Hao Su, Xuefeng Liu, Jianwei Niu, Jiahe Cui, Ji Wan, Xinghao Wu, Nana Wang
\thanks{}}

\markboth{}{}
\twocolumn[{%
\renewcommand\twocolumn[1][]{#1}%
\maketitle
\begin{center}
\vspace{-1cm}
    \centering
    \hspace*{-1.2cm}
    \includegraphics[width=8in]{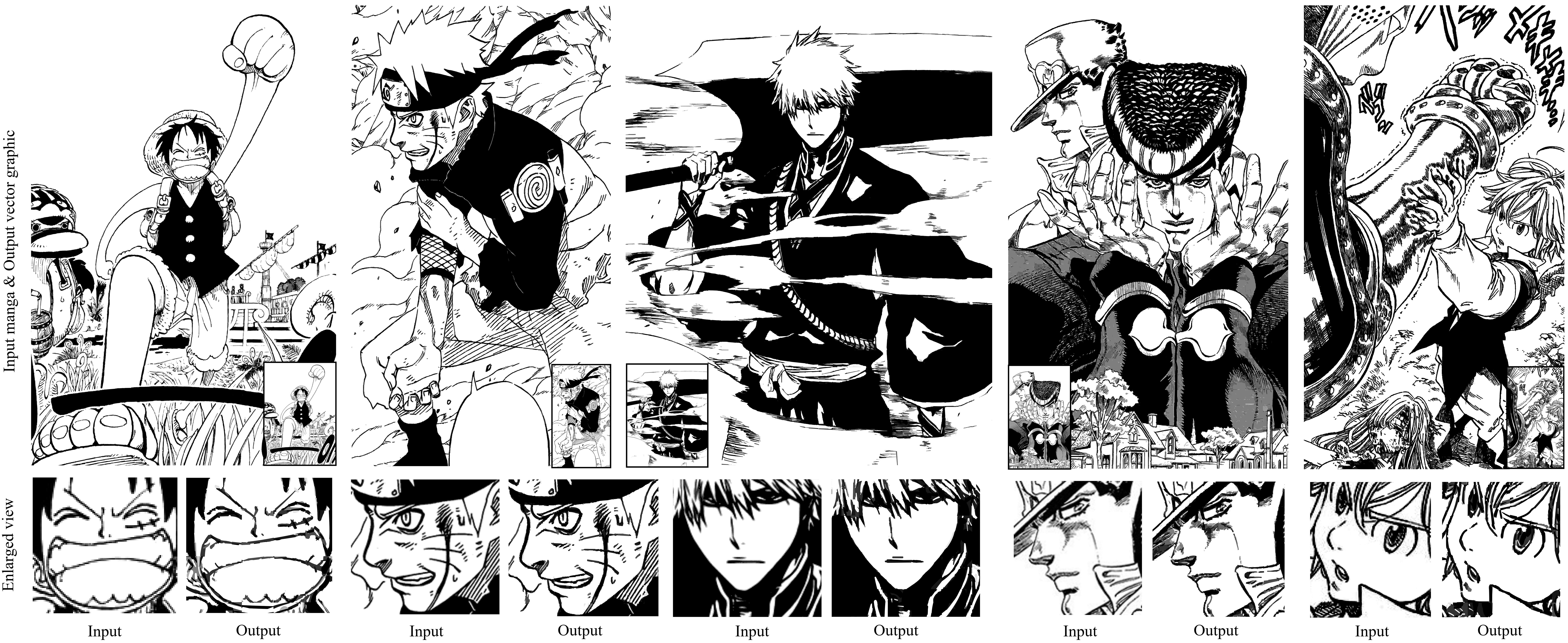}
    \vspace{-0.2cm}
    \captionof{figure}{Input raster mangas (black boxes) and our vectorized outputs. These vectorized mangas are displayed in PDF format that can be zoomed in freely.}
    \label{teaser}
\end{center}%
}]

\footnotetext{Hao Su, Jianwei Niu, Xuefeng Liu, Jiahe Cui, Ji Wan, Xinghao Wu are with State Key Lab of VR Technology and System, School of Computer Science and Engineering, Beihang University, Beijing, 100000, China. Jianwei Niu is also with Industrial Technology Research Institute, School of Information Engineering, Zhengzhou University, Henan, 450000, China; Hangzhou Innovation Institute, Beihang University, Zhejiang, 310000, China. Corresponding author: Jianwei Niu, Email: niujianwei@buaa.edu.cn.}

\begin{abstract}
Manga is a fashionable Japanese-style comic form that is composed of black-and-white strokes and is generally displayed as raster images on digital devices. Typical mangas have simple textures, wide lines, and few color gradients, which are vectorizable natures to enjoy the merits of vector graphics,  e.g., adaptive resolutions and small file sizes. In this paper, we propose MARVEL (MAnga's Raster to VEctor Learning), a primitive-wise approach for vectorizing raster mangas by Deep Reinforcement Learning (DRL). Unlike previous learning-based methods which predict vector parameters for an entire image, MARVEL introduces a new perspective that regards an entire manga as a collection of basic primitives\textemdash stroke lines, and designs a DRL model to decompose the target image into a primitive sequence for achieving accurate vectorization. To improve vectorization accuracies and decrease file sizes, we further propose a stroke accuracy reward to predict accurate stroke lines, and a pruning mechanism to avoid generating erroneous and repeated strokes. Extensive subjective and objective experiments show that our MARVEL can generate impressive results and reaches the state-of-the-art level. Our code is open-source at: \textcolor[rgb]{0.86,0.00,0.56}{https://github.com/SwordHolderSH/Mang2Vec}.
\end{abstract}


\begin{figure*}[t]
\centering
\includegraphics[width=7.1 in]{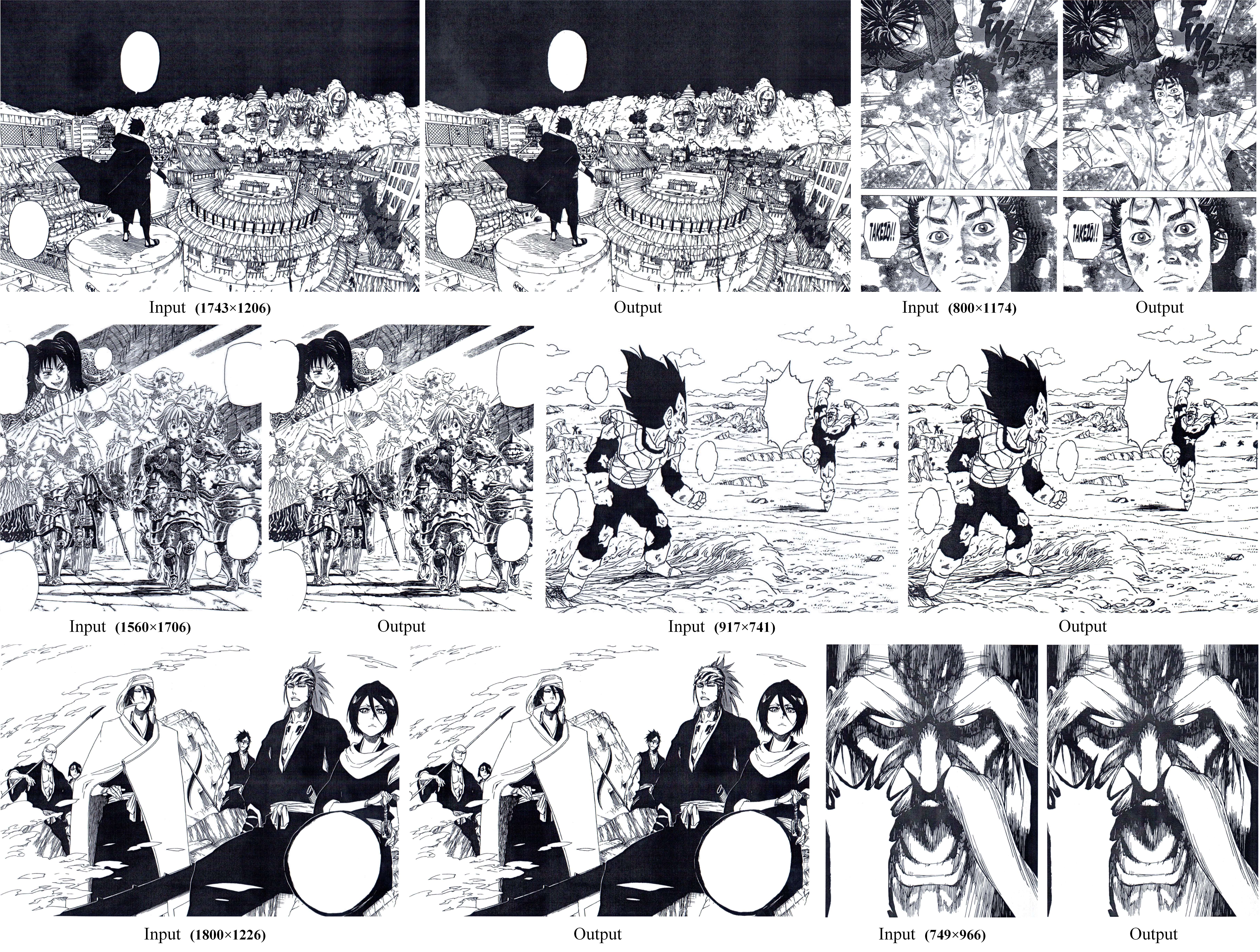}
\caption{Scan results of the print versions of raster inputs and our vectorized outputs$^1$. Our MARVEL can vectorize manga with extremely complex textures, and can effectively preserve high similarities with the raster counterparts in the print versions. The comparison in enlarged view is displayed in Figure \ref{fig:print}, and the corresponding outputs in vector formats are shown in the supplementary material.   }
\label{fig:printsample}
\vspace{-0.2cm}
\end{figure*}
\begin{figure}[t]
\centering
\includegraphics[width=3.5 in]{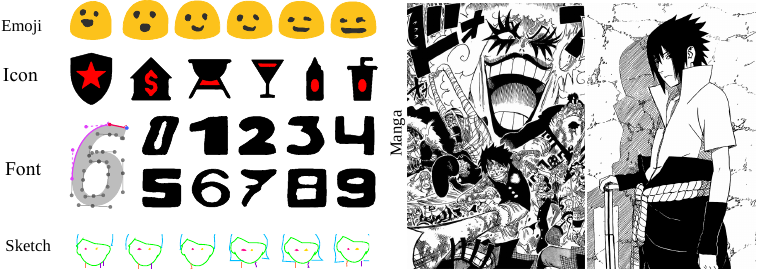}
\caption{Existing learning-based methods vectorize an entire image in a one-step manner which limits the processable vector parameters. Thus, they only have high performance in vectorizing simple textures (e.g, icons, fonts, sketches), and cannot vectorize mangas with complex structures.}
\label{fig:limit}
\vspace{-0.3cm}
\end{figure}

\section{Introduction}
\IEEEPARstart{M}{anga} is a fashionable Japanese-style comic form that is composed of black-and-white stroke lines and is generally displayed as raster images on digital devices.
As shown in Figure~\ref{teaser} and \ref{fig:printsample}, mangas have simple structures, wide lines, and few color gradients which are the potential vectorizable natures.
The main merits of vectorizing raster manga are two-fold. First, vector graphics are resolution-independent and readily displayed on digital devices with different resolutions. Second, for showing high-resolution contents, vector formats have higher compression ratios for storage than raster images.

Image vectorization has been studied extensively in vision, graphics, and other areas. Representative studies of image vectorization are divided into two categories. The first category of studies is based on pre-designed algorithms, which analyzes pixels and fits vector paths and graphics (e.g., \cite{tcsvtzhu2012object, tog2008,tvcg2009,depixelizing,line2019,tog2013cartoon, ardeco2006,Donati_2017_ICCV, dominici2020polyfit,  tcsvtding2019bilinear, tcsvtyu2019multimodal}). The other category of studies is based on deep learning (DL) (e.g., \cite{egiazarian2020deep, 2021im2vec, gao2019deepspline, sinha2017surfnet, Liu_2017_ICCV,mangagan,artcoder,su2021q, tcsvtsong2018deeply, tcsvtzhang2014unified, tcsvtbo2022all, TCSVTliu2021cross}), which trains neural models to map raster images to vector parameters.
However, existing DL-based approaches generally vectorize an entire image in one step, and the one-step manner limits the number of processable vector parameters.
Hence, as shown in Figure \ref{fig:limit}, these approaches only work well in vectorizing simple structures (e.g, fonts, icons, sketches, logos, and emojis), and cannot vectorize mangas with complex structures.

To address this issue, we present \emph{MARVEL (MAnga's Raster to VEctor Learning)}, which introduces a new perspective that decomposes an entire manga into a sequence of primitives\textemdash stroke lines for achieving accurate vectorization. 
 Employing the power of Deep Reinforcement Learning (DRL) in stepwise processing, MARVEL can predict accurate sequential stroke lines. Since the DRL model can infinitely add stroke lines to fix inaccurate areas, MARVEL outperforms related learning-based methods for vectorizing complex structures.
The merits of our approach are as follows. First, MARVEL can accurately vectorize both simple and complex textures, and can produce impressive results in  (as shown in Figure \ref{teaser} and \ref{fig:printsample}). Second, our approach only requires supervision from readily-available raster training images without high-cost vector annotation.\footnote{$^1$Each sample is printed on A4 papers (210 mm$\times$297 mm), and original resolutions of raster inputs are marked in parentheses.}

In MARVEL, a DRL agent is first learned to predict the most accurate stroke line in each timestep, where the combination of all stroke lines is constrained to follow the visual content of the target manga. Then, the predicted stroke lines are converted to vector graphics by a fitting module.
To improve our performance on vectorization accuracy and file size, we propose a stroke accuracy reward to optimize stroke prediction, and a pruning mechanism to avoid producing erroneous and repeated stroke lines.
Extensive subjective and objective experiments show that compared with both algorithm-based and learning-based methods, our MARVEL can produce impressive results and achieves state-of-the-art performance.

Our main contributions are summarized as follows:

\begin{itemize}\setlength{\itemsep}{5pt}
	\item We present MARVEL, a primitive-wise method for vectorizing raster mangas by DRL. MARVEL introduces a new perspective that considers an entire manga as a sequence of primitives\textemdash stroke lines which can be decomposed from the target image to achieve accurate vectorization.

	\item We present a new stroke accuracy reward to improve the accuracy of predicted stroke lines, and present a pruning module to avoid erroneous strokes and reduce file sizes.

    \item Experiments demonstrate that our MARVEL can generate impressive results and has high performance in vectorizing numerous types of mangas with complex structures (e.g., with intensive, sparse, wide, or thin stroke lines).

\end{itemize}

\section{Related Work}
Deep learning (DL) techniques address the single image vectorization issue utilizing the power of convolutional neural network (CNN). Below we summarize representative learning-based methods, including translative vectorization and generative vectorization.

\textbf{Translative vectorization:} 
translative methods map raster images to vector formats while preserving high visual similarities.
Egiazarian et al. \cite{egiazarian2020deep} propose a transformer-based architecture for translating technical line drawings to vector parameters. Gao et al. \cite{gao2019deepspline} produce parametric curves utilizing the extracted image features and a designed hierarchical recurrent network.
 Guo et al. \cite{guo2019deep} divide the input image into key curves by a trained CNN, and then reconstruct the topology at junctions by predicting the line connectivity. For vectorizing floorplans, Liu et al. \cite{Liu_2017_ICCV} first train a CNN to convert an image to junctions, and then address an integer program that obtains the vectorized floorplans as a set of architectural primitives. For line-art images, Mo et al. \cite{mo2021general} propose a framework of recurrent neural network that can produce vectorized line drawings.
 Reddy et al. \cite{2021im2vec} propose Im2Vec, a method based on variational autoencoder to predict vectorization parameters of input images, and the network can be trained without vector supervision. Moreover, there is a special method LIVE \cite{ma2022towards} that is not a typical learning-based method, which fits vector curves iteratively using a neural optimizer without training any model. LIVE requires hundreds of iterations to fit each vector curve and spends more time than vectorization methods using a trained model.

\begin{figure*}[t]
\centering
\vspace{0.5cm}
\includegraphics[width=7 in]{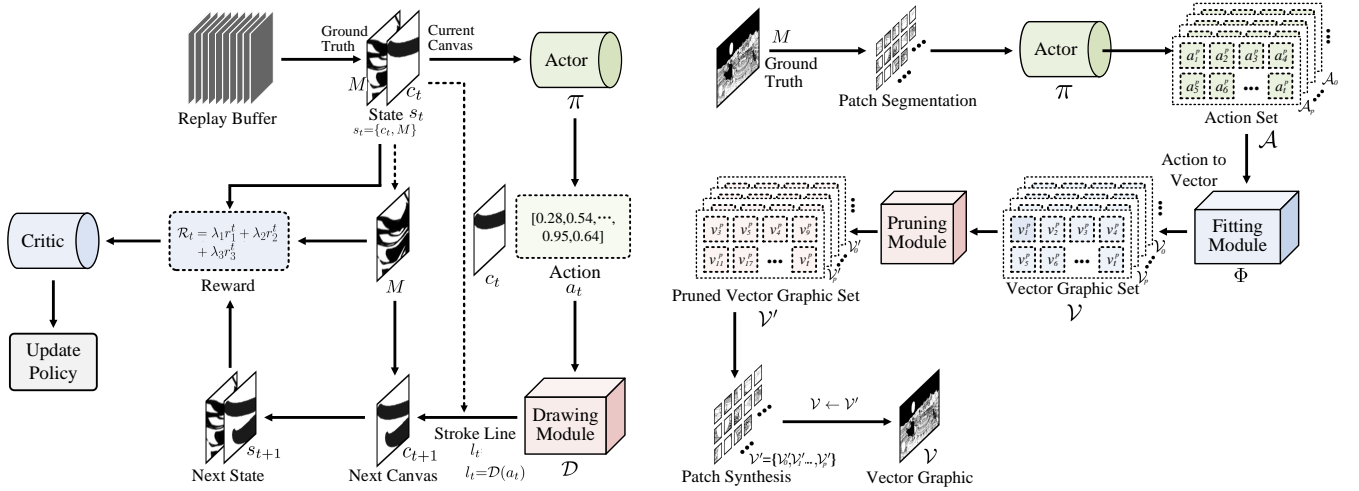}
\caption{System Pipeline. Given a raster manga ${M}$, our method maps $M$ to a vector graphic ${\mathcal{V}}$ with high visual similarity.~\textbf{Left:} the learning phase $\Psi_l$ aims to learn a DRL model to predict an action sequence $\mathcal{A}=\Psi_l(M)$, and $\mathcal{A}$ can be rendered to a stroke line sequence $\mathcal{L}$ which is encouraged to compose $M$ visually.~\textbf{Right:} the vectorization phase $\Psi_v$ aims to translate action sequence $\mathcal{A}$ to vector format $\mathcal{V}$.}
\label{fig:overview}
\end{figure*}

\textbf{Generative vectorization:} the goal of generative models is to predict vector parameters by some heuristic inputs (e.g, sketches, texts, numbers, conditional parameters), where no accurate similarities are required between inputs and outputs.
SketchRNN \cite{ha2017neural} encodes the input sketches to pen positions and states, and a recurrent neural network is trained to predict a new sketch. Similar to SketchRNN, Sketchformer \cite{ribeiro2020sketchformer} adopts a framework to encode vector form sketches using the transformer.
SVG-VAE \cite{svgvae} fixes the weights of pre-trained variational auto encoder weights and trains a decoder to predict vector parameters from the latent variable. DeepSVG \cite{deepsvg} shows that the hierarchical networks are useful to reconstruct diverse vector graphics, and do well in interpolation and generation tasks. For font glyphs vectorization, methods \cite{azadi2018multi, gao2019artistic} can produce results from partial observations in a low-resolution raster domain. Li et al. \cite{differentiable} present a differentiable rasterizer to edit and produce vector parameters by raster-based target functions and machine learning.

Existing learning-based methods focus specifically on vectorizing simple contents (e.g., fonts, numbers, sketches, line-drawings), since their neural models vectorize an entire image in a one-step manner which limits the number of processable vector parameters.
Unlike these methods, our MARVEL considers a complex image as a combination of vector primitives\textemdash stroke lines, and produces accurate primitives using the DRL's merit, i.e., predicting vector parameters in a stepwise manner.

There are also some studies that involve the simplification or beautification \cite{tog2017r1,ICCV2021r2,TOG2021r3,cvpr2021r4,tog2020r5, tog2018r6, r7han,r8liu2015closure,r9qu2008richness,r10wang2006deringing} of line-art (e.g., manga, sketch), which is a crucial step for migrating raster images to the vector domain.
Amit Shesh and Baoquan Chen \cite{shesh2008efficient} propose an approach for simplifying 2D/3D line drawings by creating and managing a time-coherent hierarchy. Smart Scribbles \cite{noris2012smart} introduce a scribble-based interface for user-guided segmentation of sketchy drawings.
Liu et al. \cite{r8liu2015closure} propose a novel approach to simplify sketch drawings, which makes the first attempt in incorporating the law of closure into the semantic analysis of the sketches.
For simplifying manga drawing, Li et al. \cite{tog2017r1} propose a tailored deep network framework for extracting structural lines from mangas with arbitrary screen patterns.


\section{Method}
\subsection{Overview}
Given a raster manga ${M}$, our MARVEL is modeled as a function $ {\Psi}$ to generate a vector graphic $\mathcal{V}={\Psi}({M})$, and $\mathcal{V}$ is similar to $M$ visually.
As shown in Figure~\ref{fig:overview}, MARVEL $\Psi$ is composed of two phases, the learning phase $\Psi_l$ (Figure~\ref{fig:overview} left) and the vectorization phase $\Psi_v$ (Figure~\ref{fig:overview} right). $\Psi_l$ is designed to learn a DRL model for decomposing $M$ to a sequence of stroke lines. Specifically, following the visual content of $M$, a DRL agent is trained to predict an action sequence ${\mathcal{A}}=\Psi_l(M)= \{$$a_0,a_1...,a_t$$\}$, and $\mathcal{A}$ can be rendered to a sequence of stroke lines $\mathcal{L}$$=$$\{$$l_0,l_1...,l_t$$\}$ by a drawing model $\mathcal{D}$, $l_t$$=$$\mathcal{D}(a_t)$. Meanwhile, the combination of strokes in $\mathcal{L}$ are constrained to compose $M$ visually.
$\Psi_v$ is designed to translate ${\mathcal{A}}$ to the vector form $\mathcal{V}$. That is, leveraging the well-trained DRL agent, we predict the action sequence $\mathcal{A}$, and then fit $\mathcal{A}$ to vector graphic $\mathcal{V}$ by a fitting module.

 We will detail $\Psi_l$ and $\Psi_v$ in Section \ref{sec:learn} and Section \ref{sec:vectorization} respectively. Table I summarizes the key notations used throughout the paper.

\renewcommand\arraystretch{1.3}
\begin{table}[t] \small
	\caption{Summary of Notations}
	\centering
	\begin{tabular}{p{0.7cm}<{\centering}|p{7cm}}
		\hline
		\textbf{Name} & \textbf{Description}   \\
		\hline
		$M$   &  The input raster manga (ground truth). \\
		$\mathcal{V}$  & The output vector graphic by vectorizing $M$, consisting of a sequence of vectorized strokes, $\mathcal{V}$$=$$\{$$v_0,v_1...,v_t$$\}$.\\
		$\mathcal{A}$  & The sequence of actions, $\mathcal{A}$$=$$\{$$a_0,a_1...,a_t$$\}$. \\
        $\mathcal{L}$  & The sequence of stroke lines, $\mathcal{L}$$=$$\{$$l_0,l_1...,l_t$$\}$.  \\
        $\mathcal{D}$  & The drawn module used to render $a_{t}$ to $l_{t}$, $l_t$$=$$\mathcal{D}(a_t)$.  \\
		$a_{t}$  & The $t$-th action predicted by agent $\pi$. \\	
		$l_{t}$  & The $t$-th stroke line translated by $a_t$. \\
        $v_{t}$  & The $t$-th vectorized stroke line corresponding to $l_{t}$. \\
        $c_{t}$  &The $t$-th canvas produce by rendering stroke lines, $c_{t+1}=c_t+l_t$. The initial canvas $c_0$ is blank. \\
        $s_{t}$  &The $t$-th state that contains $c_t$ and $M$, $s_t$$=$$(c_t,M)$. \\
        $p$  &The number of patches. \\
        $k$  &The maximum number of stroke lines, i.e., the maximum timestep of the DRL model. \\
		\hline
	\end{tabular}
	\label{table:meaning}
\end{table}
\renewcommand\arraystretch{1}

\subsection{Learning Phase}
\label{sec:learn}
The goal of $\Psi_l$ is to train a DRL agent $\pi$ to predict a optimal action sequence $\mathcal{A}$$=$$\{$$a_0,a_1...,a_t$$\}$. In each timestep $t$, an action $a_t$$=$$\pi(s_t)$ is predicted by the observed current state $s_t$, and $a_t$ can be rendered to a stroke line $l_t=\mathcal{D}(a_t)$ by the drawing module $\mathcal{D}$. All stroke lines are rendered on a blank canvas sequentially, and the rendered canvas is constrained to be similar to $M$.

 To improve the accuracies of stroke lines, our learning strategy follows the \emph{Greedy Strategy} and \emph{Markov Decision Process}. Specifically, regardless of the previous and future canvas, we only focus on the current canvas and predict the optimal stroke line which makes the next canvas most similar to $M$.

\textbf{Basic~of~model-based~DRL:}~our MARVEL roughly utilizes the framework of model-based DRL. Compared with model-free DRL methods that sample at will from environments, model-based DRL methods construct dynamics models to simulate the environments as they sample \cite{modelbasedDRL}. Leveraging dynamics models for policy updates, the complexity and difficulty of sampling can be reduced significantly. The key to a successful model-based DRL method is to design a suitable policy, state, reward, and dynamics model. Below we detail the design of MARVEL.

\textbf{Policy:} as shown in Figure~\ref{fig:overview} left, our method follows the policy of model-based Deep Deterministic Policy Gradient (DDPG) \cite{ddpg} that uses the actor-critic architecture \cite{konda2000actor}. Our architecture consists of two networks, the actor $\pi(s_t)$ and critic $Q(s_t)$. The actor models a policy $\pi$ that predicts action $a_t$ from state $s_t$, and the critic estimates the current reward $\mathcal{R}_t$ by
\begin{equation}
\label{equation:Q}
Q(s_t)= \mathcal{R}_t(a_t, s_t) + \gamma Q(s_{t+1}),
\end{equation}
where $\mathcal{R}_t(a_t, s_t)$ is the current reward calculated by $a_t$ and $s_t$, $\gamma$ is the discount factor, and the actor $\pi(s_t)$ is trained to maximize $Q(s_t)$ estimated by the critic.

\textbf{State:} our defined state $s_t$ is independent from timestep $t$, and $s_t$ consists of two parts: the current canvas $c_t$ and the target manga $M$, represented as $s_t$$=$$\{c_t,M\}$. Initializing with a blank canvas $c_0$, the actor $\pi$ predicts action $a_t$$=$$\pi(s_t)$ according to $t$-th state $s_t$, and the ($t$$+$$1$)-th canvas $c_{t+1}$$=$$c_{t}$$+$$\mathcal{D}(a_t)$ is rendered by the drawing module $\mathcal{D}$, $c_t$$=$$c_0$$+$$\sum_{k\in\{0,1,...,t\}}l_k$. Then, the next state $s_{t+1}$ is represented as $s_{t+1} = \{ \ \! c_t$$+$$\mathcal{D}(\pi(s_t)), M \ \! \}$.

\begin{figure}[t]
\vspace{0.5cm}
\centering
\includegraphics[width=3.2 in]{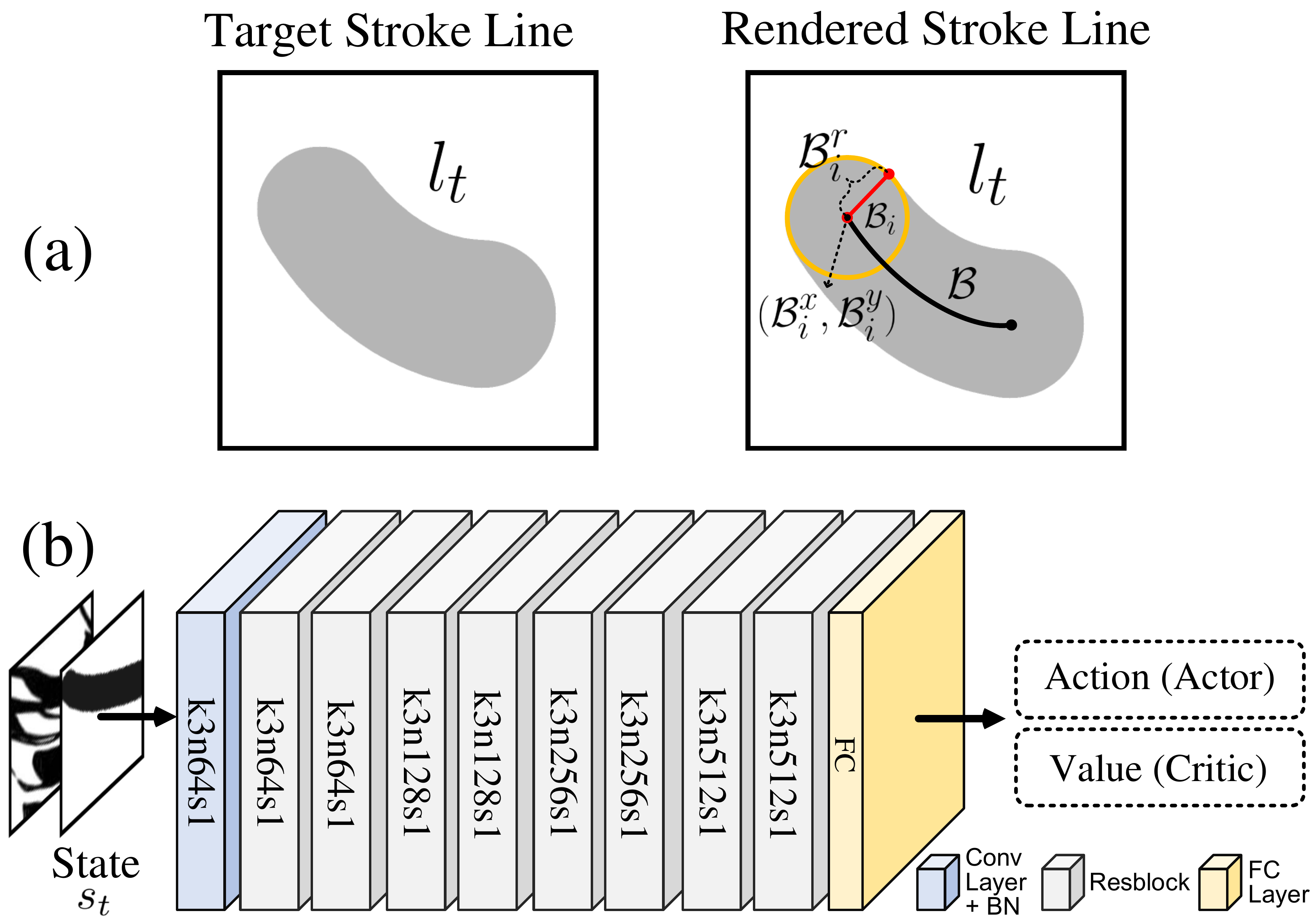}
\caption{(a) The detail of drawing module $\mathcal{D}$ to render a stroke line $l_t$. (b) Network architectures of actor and critic.}
\label{fig:network}
\end{figure}

\textbf{Drawing module:} we design a drawing module $\mathcal{D}$ as the dynamics model of our model-based DRL, and $\mathcal{D}$ is employed to render an action $a_t$ to a stroke line $l_t$, formulated as $l_t$$=$$\mathcal{D}(a_t)$, $a_t$$\in$$[0,1]$. Although training a neural render to map $a_t$ to $l_t$ is flexible (e.g., \cite{l2p}), it compromises stroke lines' accuracies (e.g., irregular or blurring edges, losing pixels). Therefore, $\mathcal{D}$ is design by an image rendering algorithm.

As shown in Figure \ref{fig:network}(a), the path of $l_t$ is designed to follow the rules of Quadratic B\'{e}zier Curve (QBC), and the $i$-th point $\mathcal{B}_i$ on a QBC $\mathcal{B}$ is represented as
\begin{equation}
\label{equation:QBC}
\begin{aligned}
\mathcal{B}_i^x & = (1 - i)^{2}P_0^x + 2(1 - i)iP_1^x + i^{2}P_2^x  \\
\mathcal{B}_i^y & = (1 - i)^{2}P_0^y + 2(1 - i)iP_1^y + i^{2}P_2^y  \\
\mathcal{B}_i^r & = (1 - i) P_0^r + i P_2^r
\end{aligned} ,
\end{equation}
where ($\mathcal{B}_i^x$, $\mathcal{B}_i^y$) and $\mathcal{B}_i^r$ indicate the ($x$, $y$) coordinates and the radius of $\mathcal{B}_i$ respectively, $i\in[0,1]$. $P_0$, $P_1$, and $P_2$ are the three control points of QBC $\mathcal{B}$.

 \textbf{Action:} corresponding to $\mathcal{D}$, the designed action $a_t$ consists of nine control parameters of a QBC, defined as
 \begin{equation}
\label{equation:action}
\begin{aligned}
a_t = ({P_0^x}, \ \!  P_0^y,  \ \! P_1^x,  \ \! P_1^y,  \ \! P_2^x,  \ \! P_2^y,  \ \! P_0^r,  \ \! P_2^r,  \ \! g)_t
\end{aligned} ,
\end{equation}
where $(P_0^x, P_0^y)$, $(P_1^x, P_1^y)$, and $(P_2^x, P_2^y)$ indicate the $(x,y)$ coordinates of $P_0$, $P_1$, and $P_2$ respectively. $P_0^r$ and $P_2^r$ control the width of a stroke line, and $g$ controls the color in the 1-channel grayscale space, containing 11 gray levels, i.e., $255\times\{0,0.1,0.2,...,1\}$.

\begin{figure}[t]
\vspace{0.5cm}
\centering
\includegraphics[width=3.5 in]{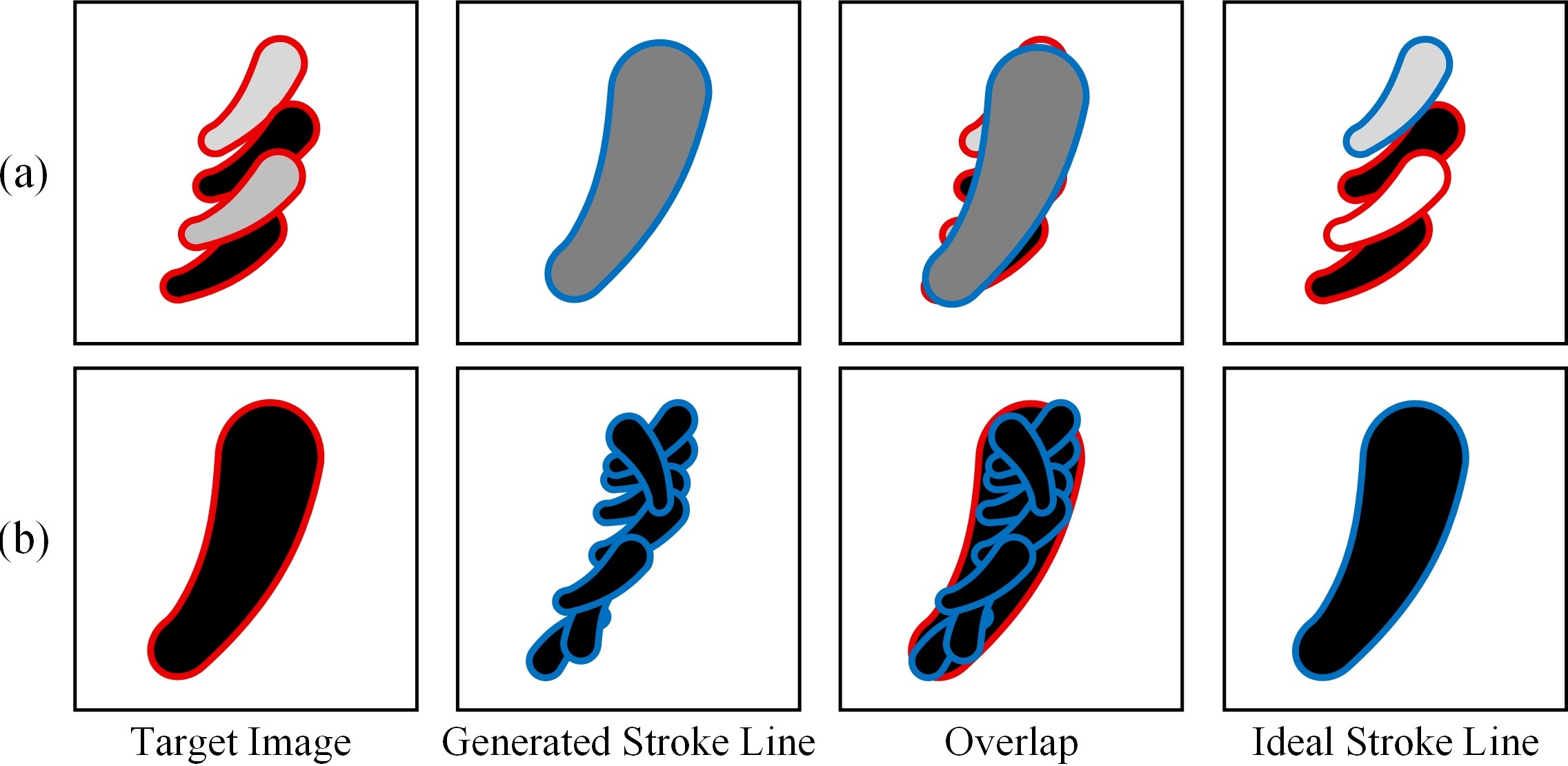}
\caption{The baseline reward $\mathcal{R}_{L_2}$ incurs the trained agent to predict inaccurate stroke lines. (a) For a target with strokes in different colors, the agent generates a larger stroke with an average color. (b) For a target with a larger stroke, the agent generates numerous smaller strokes repeatedly. Although these two situations also satisfy the increase of $\mathcal{R}_{L_2}$ , the predicted stroke lines are inaccurate and undesired.}
\label{fig:inaccurate}
\end{figure}

\textbf{Reward:} the basic idea of our designed reward is that in each timestep $t$, we encourage $\pi$ to predict the most accurate stroke line $l_t$ which makes the next canvas most similar to $M$. Huang \emph{et al.} \cite{l2p} propose that employing the L2 reward $\mathcal{R}_{L_2}$ can train the agent to draw a target image gradually, represented as
 \begin{equation}
\label{equation:rl2}
\begin{aligned}
\mathcal{R}_{L_2} & = \| c_t - M \|^2 - \| c_{t+1} - M \|^2\\
\end{aligned} \ \ ,
\end{equation}
where $\mathcal{R}_{L_2}$ means to encourage the next canvas $c_{t+1}$ to be more similar to $M$ than the current canvas $c_{t}$.

However, we observe in experiments that $\mathcal{R}_{L_2}$ typically incur the agent to predict inaccurate stroke lines.
First, as shown in Figure~\ref{fig:inaccurate}(a), for targets with strokes in different colors, the agent produces a larger stroke with an average color. Second, as shown in Figure~\ref{fig:inaccurate}(b), for targets with a larger stroke line, the agent will produce numerous smaller strokes repeatedly. Although these two situations also satisfy the increase of $\mathcal{R}_{L_2}$, the predicted stroke lines are inaccurate and unideal.

To predict accurate stroke lines, we propose a \emph{Stroke Accuracy} reward $\mathcal{R}_{t} $. 
Let $c_t$, $c_{t+1}$, and $l_t$ indicate the current canvas, the next canvas, and the stroke line respectively, $c_{t+1} = c_{t} + l_t$. We divide $l_t$ into two parts, the stroke area $l_t^b$ and the stroke color $g_t\in[0, 1]$, $l_t = l_t^b \times g_t$. $l_t$, $l_t^b$, and $c_t$ have the same shape of $C \! \times\! H \! \times W$. $l_t^b$ is a matrix consisting of 1 and 0, where each entry in the stroke area is set to 1, and other entries are set to 0. Then, our reward $\mathcal{R}_{t} $ is defined as:
\begin{equation}
\label{equation:r}
\begin{aligned}
&\mathcal{R}_t  = \lambda_1 r^t_1+  \lambda_2 r^t_2+ \lambda_3 r^t_3  \\
&r^t_1(l_t)  = \frac{1}{C H W}\cdot \!\!\!\!\!  \sum_{(i,j)\in l_t^b} \!\!\! l_t^b(i,j) \\
&r^t_2(l_t, c_t, c_{t+1})  =   \frac{1}{C H W} \big\| l_t^b \! \times \! c_t  - l_t^b \! \times \! c_{t+1} \big\|^2_2 \\
&r^t_3(l_t,c_{t+1},M)  =  1- \frac{1}{C H W} \big\|l_t^b \! \times\! c_{t+1}  - l_t^b \! \times \! M \big\|^2_2
\end{aligned} ,
\end{equation}
where $\lambda_1$ to $\lambda_3$ are used to balance the multiple objectives, $(i,j)$ is an entry in matrix $l_t^b$, and $\{r^t_1$, $r^t_2$, $r^t_3\}$$\in$$[0,1]$. The underlying ideas of the rewards are as follows:
\begin{itemize}
	\item $r^t_1$ encourages $\pi$ to maximize the area of generated stroke line $l_t$.
    \item $r^t_2$ encourages $\pi$ to maximize the difference between $l_t^b$$\times$$c_t$ and $l_t^b$$\times$$c_{t+1}$.
    \item $r^t_3$ encourages $\pi$ to maximize the similarity between $l_t^b$$\times$$c_t$ and $l_t^b$$\times$$M$.
\end{itemize}



\begin{figure}[t]
\centering
\includegraphics[width=3.3 in]{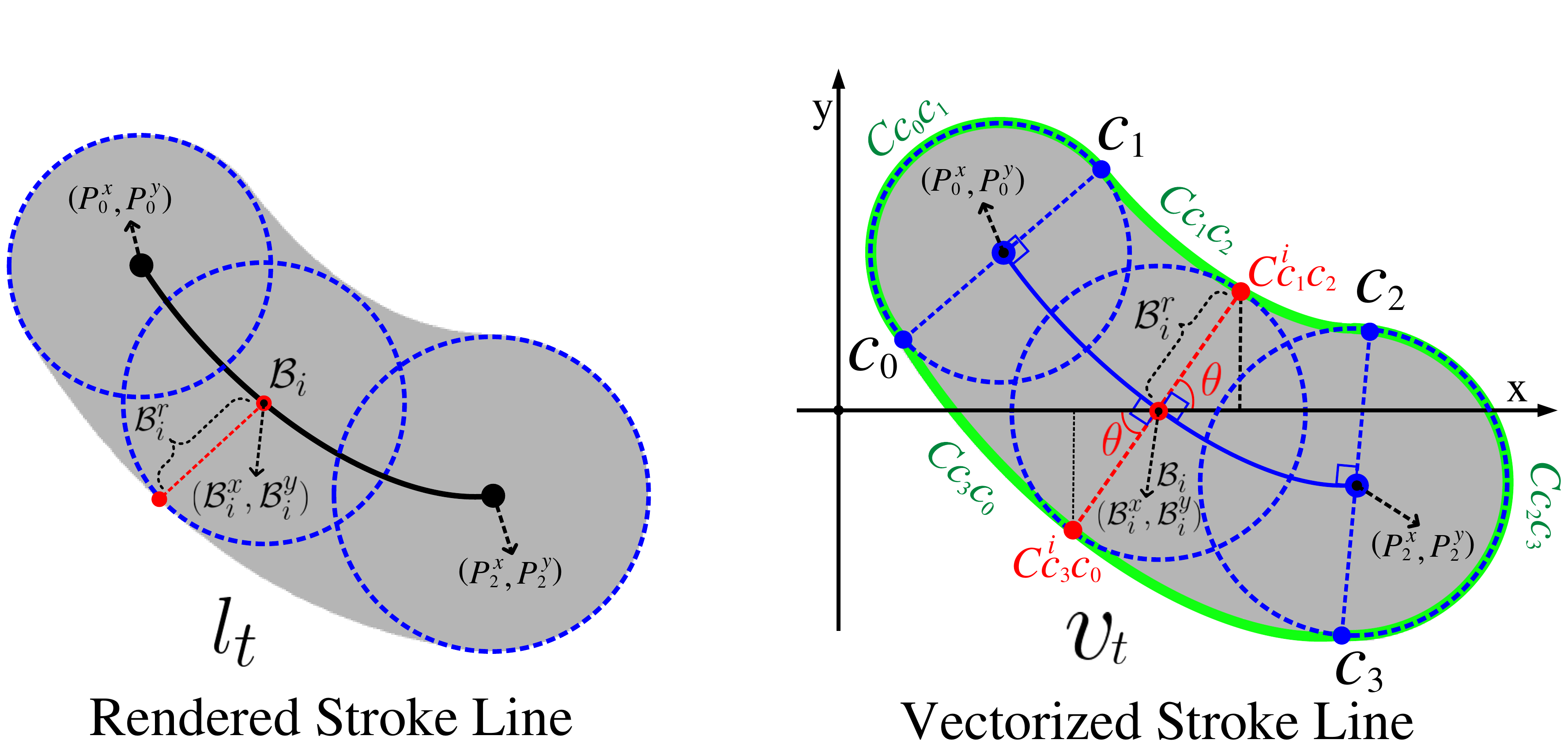}
\caption{Design of vectorized stroke line $v_t$. \textbf{Left:} the rendered stroke line $l_t$. \textbf{Right:} the corresponding vectorzied stroke line $v_t$, where $v_t$ and $l_t$ have the same geometry. The key to producing $v_t$ is to fit the QBC representation of the green path $C$ which consists of four curves $C_{c_0c_1}$, $C_{c_1c_2}$, $C_{c_2c_3}$, and $C_{c_3c_0}$.}
\label{fig:fiting-module}
\end{figure}

\textbf{Network architecture:}
as shown in Figure \ref{fig:network}(b), to extract features of manga textures with high complexity, our actor and critic use the residual structure similar to ResNet-18 \cite{resnet}. The actor utilizes the Batch Normalization \cite{BN} and the critic utilizes Weight Normalizationn \cite{wn} with Translated ReLU (TReLU) [31] to stabilize the learning, and all input training images are resized to $128\times128$.
According to the method of model-based DDPG \cite{l2p}, we utilize the soft target network that constructs a copy for the actor and critic, and updates their weights by making them slowly track the learned networks.

\subsection{Vectorization Phase}
\label{sec:vectorization}
Figure~\ref{fig:overview}~right shows the pipeline of vectorization phase $\Psi_v$. First, we divide raster manga $M$ into $p$ patches $M$$=$$\{M_0,M_1...,M_p\}$ to adapt to the input size of actor $\pi$. Next, utilizing the well-trained $\pi$, we predicts an action sequence ${\mathcal{A}_p}$$=$$\pi(M_p)$$=$$\{a^p_0,a^p_1,...,a^p_t\}$ for each patch $p$. Then, $\mathcal{A}_p$ is fitted to a sequence of vectorized stroke lines $\mathcal{V}_p$$=$$\Phi(\mathcal{A}_p)$ by a fitting module $\Phi$. Finally, the pruning module optimizes $\mathcal{V}_p$'s file size and accuracy to output the vectorization result $\mathcal{V}=\{\mathcal{V}_0,\mathcal{V}_1,...,\mathcal{V}_p\}$. 

\begin{algorithm}[t]
	\caption{Pruning Algorithm.}
    \setstretch{0.9}
	\LinesNumbered
	\KwIn{$\mathcal{V}$, $M$, $\xi $;}
     Initialize $\mathcal{V}'= \varnothing$ \;
    \For {$\mathcal{V}_p \in  \mathcal{V}$}
    {
    	$k \leftarrow |\mathcal{V}_p|$  \;
        $t \leftarrow k$  \;
        $\delta \leftarrow \frac{\|M_p- R(\mathcal{V}_p)\|^2_2}{C  H  W}$ \;
    	\While {$t>0 $}
    	{
          $\mathcal{V}'_p \leftarrow \mathcal{V}_p$ \;
          Remove $v_t^p$ in $\mathcal{V}_p$\;
          $\delta' \leftarrow \frac{\|M_p- R(\mathcal{V}_p)\|^2_2}{C  H  W}$ \;
             \uIf{$\delta' \! \leqslant \delta+\xi $}
             { $\delta \leftarrow \delta' $  \;
               $\mathcal{V}'_p \leftarrow \mathcal{V}_p$  \;
               $t \leftarrow t-1$ \;
             }
             \Else
             {
               $\mathcal{V}_p \leftarrow \mathcal{V}'_p$  \;
              $t \leftarrow t-1$ \;
             }
    	}
    Insert $\mathcal{V}'_p$ into $\mathcal{V}'$ \;
    }
\KwOut{$\mathcal{V}'$;}
\end{algorithm}

\textbf{Action to vectorized stroke line}: in fitting module $\Phi$, each action $a_t$ is translated to a vectorized stroke $v_t=\Phi(a_t)$, where $v_t$ and $l_t$ have the same geometry. As shown in Figure \ref{fig:fiting-module} right, the key to producing $v_t$ is to fit the QBC representation of the green path $C$ which is composing of four curves $C_{c_0c_1}$, $C_{c_1c_2}$, $C_{c_2c_3}$, and $C_{c_3c_0}$.
First, following the definition of $a_t$ in Eq.(\ref{equation:action}), $C_{c_0c_1}$ and $C_{c_2c_3}$ can be easily obtained by fitting the arcs of circles whose center coordinates are $(P_0^x, P_0^y)$ and $(P_2^x, P_2^y)$ respectively. Then, let $C_{c_1c_2}^i$ indicates the $i$-th point on $C_{c_1c_2}$, the $x$ and $y$ coordinates of $C_{c_1c_2}^i$ are computed by
\begin{equation}
\label{equation:fit}
\begin{aligned}
C_{c_1c_2}^{i,x} = \mathcal{B}_i^x + \mathrm{cos}(\theta) \! \cdot \! \mathcal{B}_i^r , \ \ C_{c_1c_2}^{i,y} = \mathcal{B}_i^y + \mathrm{sin}(\theta) \! \cdot \! \mathcal{B}_i^r \ ,
\end{aligned}
\end{equation}
where $\mathcal{B}_i^x$, $\mathcal{B}_i^y$ and $\mathcal{B}_i^r$ are defined in Eq.(\ref{equation:QBC}), $\theta$$=$$\mathrm{arctan}[(\mathcal{B}_i')^{-1}]$, and $\mathcal{B}_i'$ indicates the derivative of $\mathcal{B}_i$. Combining Eq.(\ref{equation:QBC})(\ref{equation:action}), $\mathcal{B}_i'$ is calculated by
\begin{equation}
\label{equation:fit2}
\begin{aligned}
\mathcal{B}'_i & = \frac{\mathrm{d}\mathcal{B}_i^y}{\mathrm{d}\mathcal{B}_i^x}= \frac{(1-i)(P_1^y-P_0^y)+i(P_2^y-P_1^y)}{(1-i)(P_1^x-P_0^x)+i(P_2^x-P_1^x)} \\
\end{aligned}.
\end{equation}
Following Eq.(\ref{equation:fit}) and (\ref{equation:fit2}), though fitting the computable points on $C_{c_1c_2}$ and $C_{c_3c_0}$, we can obtain the QBC representations of each curves, and finally fit the path $C$ and the vectorized stroke $v_t$.

\begin{figure*}[t]
\centering
\vspace{-0.1cm}
\includegraphics[width=7.2 in]{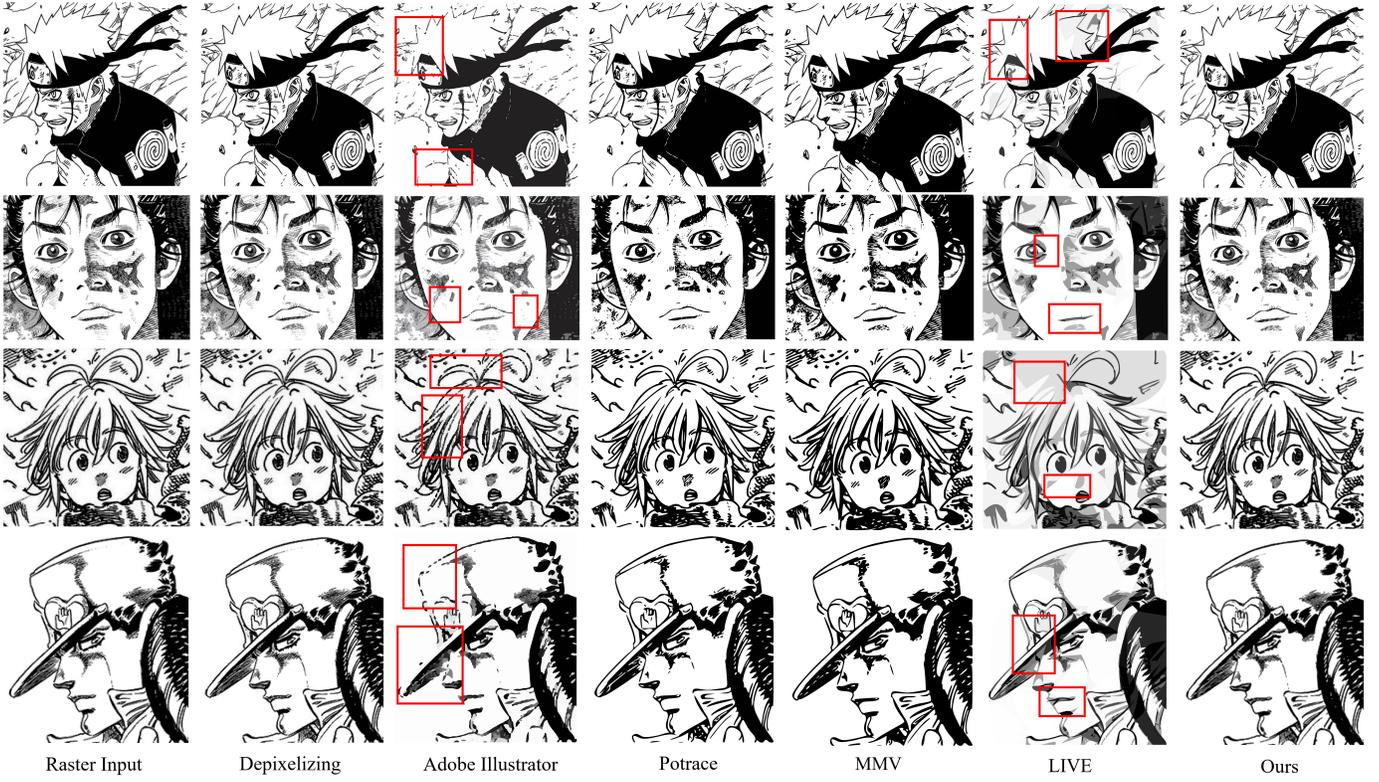}
\caption{Comparison with representative methods and commercial packages that can vectorize complex contents, including Depixelizing \cite{depixelizing}, Adobe Illustrator \cite{adobe}, Potrace \cite{potrace}, MMV \cite{TVCG2016manga}, and LIVE \cite{ma2022towards}.   }
\label{fig:compare-algorithm-based}
\end{figure*}

\textbf{Pruning module}: in $\Psi_v$, there are two issues compromise the performance on vectorization. First, the actor $\pi$ may produce erroneous strokes that reduce the vectorization accuracy. Second, some repeated or redundant strokes increase the file sizes of vectorized results.

To address these two issues, we propose a pruning module to optimize the erroneous and repeated stroke lines.
As shown in Algorithm 1, we input the vectorized result $\mathcal{V}$ of $p$ patches and output the pruned result $\mathcal{V}'$, where $\mathcal{V}$$=$$\{\mathcal{V}_0,\mathcal{V}_1...,\mathcal{V}_p\}$ and $\mathcal{V}_p=\{v_0^p,v_1^p...,v_t^p\}$. $M$ is the target raster manga, and $\xi$ indicates the tolerable error that is used to trade-off the visual similarity and the file size of $\mathcal{V}'$. In the 3-rd line, $|\mathcal{V}_p|$ indicates the cardinality of $\mathcal{V}_p$ (i.e., the maximum number of stroke lines). In the 5-th line, the function $R(\mathcal{V})$ maps $\mathcal{V}$ to a raster image by CairoSVG \cite{cairosvg}, and $\delta$ is the measured difference between $R(\mathcal{V}_p)$ and $M_p$. $M_p$ and $R(\mathcal{V}_p)$ have the same shape of $C$$\times$$H$$\times$$W$.

\begin{figure}[t]
\centering
\includegraphics[width=3.5 in]{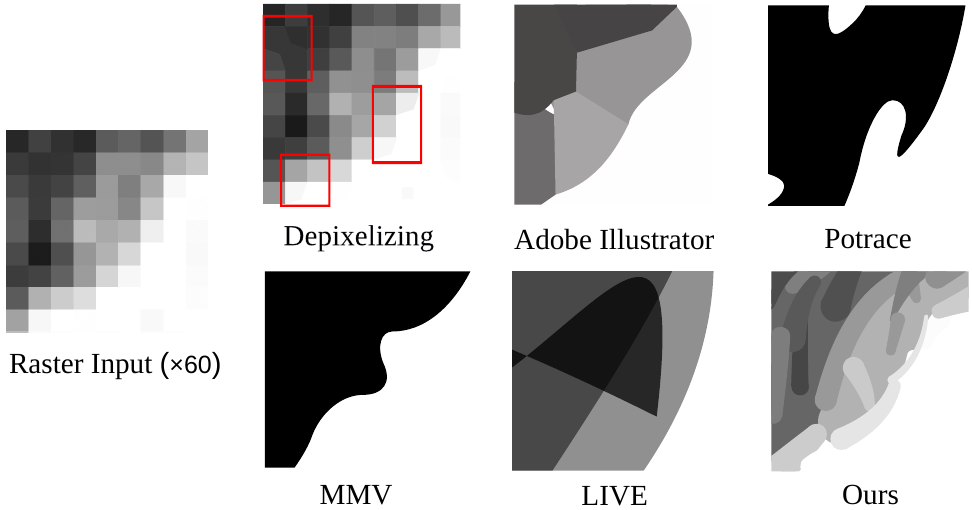}
\caption{Enlarged view ($\times$60) of samples in Figure \ref{fig:compare-algorithm-based}. Although Depixelizing \cite{depixelizing} generates accurate vectorized results, the method converts all pixels to square vector grids and only smooths several critical paths, which incurs the results to look as blurry as the raster inputs (red boxes). By contrast, our method achieves a better balance in accuracy and smoothness.} 
\label{fig:compare-algorithm-based-large}
\end{figure}

\section{Experiment}
Below we evaluate the performance of MARVEL in 5 aspects: vectorization accuracy, visual effect, file sizes, time cost, and print quality.
\subsection{Implementation}
\textbf{Dataset:} for training MARVEL, we collect a dataset \emph{DeepManga} from several popular manga works. DeepManga contains 42599 raster mangas with different resolutions. In the learning phase, each training data is converted to 1-channel grayscale space, and we randomly cut 128$\times$128 images from original data as the inputs.

\textbf{Experimental setting:} our MARVEL is implemented in PyTorch, and all comparison methods and experiments are performed on a computer with an NVIDIA Geforce RTX 2080 GPU, and 16 Intel(R) i7-10700F CPU. The learning rates of actor (or critic) are from $3$$\times$$10^{-4}$ to $1$$\times$$10^{-4}$ (or $1$$\times$$10^{-3}$ to $3$$\times$$10^{-4}$), which decays after $1$$\times$$10^5$ training batches. We set 1 action per timestep, 40 timesteps per episode, and the reward discount factor is $0.95^5$. In all experiments, by default, we set $\lambda_1$$,$$\lambda_2$$,$$\lambda_3=1$ in Eq.(\ref{equation:r}), the maximum number of strokes $k=40$, and the number of patches $p=32\times32$.

\textbf{Evaluation~index~of~vectorization~accuracy:} for evaluating the vectorization accuracy objectively, we leverage indexes \emph{MSE (Mean Square Error)} and \emph{SSIM} \cite{ssim} \emph{(Structural SIMilarity)} to measure the similarities between inputs and outputs. In evaluations, each result in vector space is rasterized to images of size 512$\times$512 by CairoSVG \cite{cairosvg}. The MSE index is computed by $\frac{\|I - O\|^2_2}{CHW \times 255}$$\in$$[0,1]$, where $I$ and $O$ indicate the raster input and the vectorized output of shape $C$$\times$$ H$$\times$$W$. The SSIM index is in $[-1, 1]$, and the index value is proportional to the similarity between two images.

\begin{figure}[t]
\centering
\includegraphics[width=3.5 in]{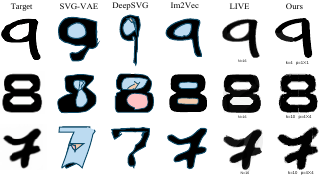}
\caption{Comparison with learning-based methods in vectorizing images with simple structures (e.g., F\scriptsize{ONTS}\normalsize{}).}
\label{fig:compare_s}
\end{figure}

\begin{figure}[t]
\centering
\includegraphics[width=3 in]{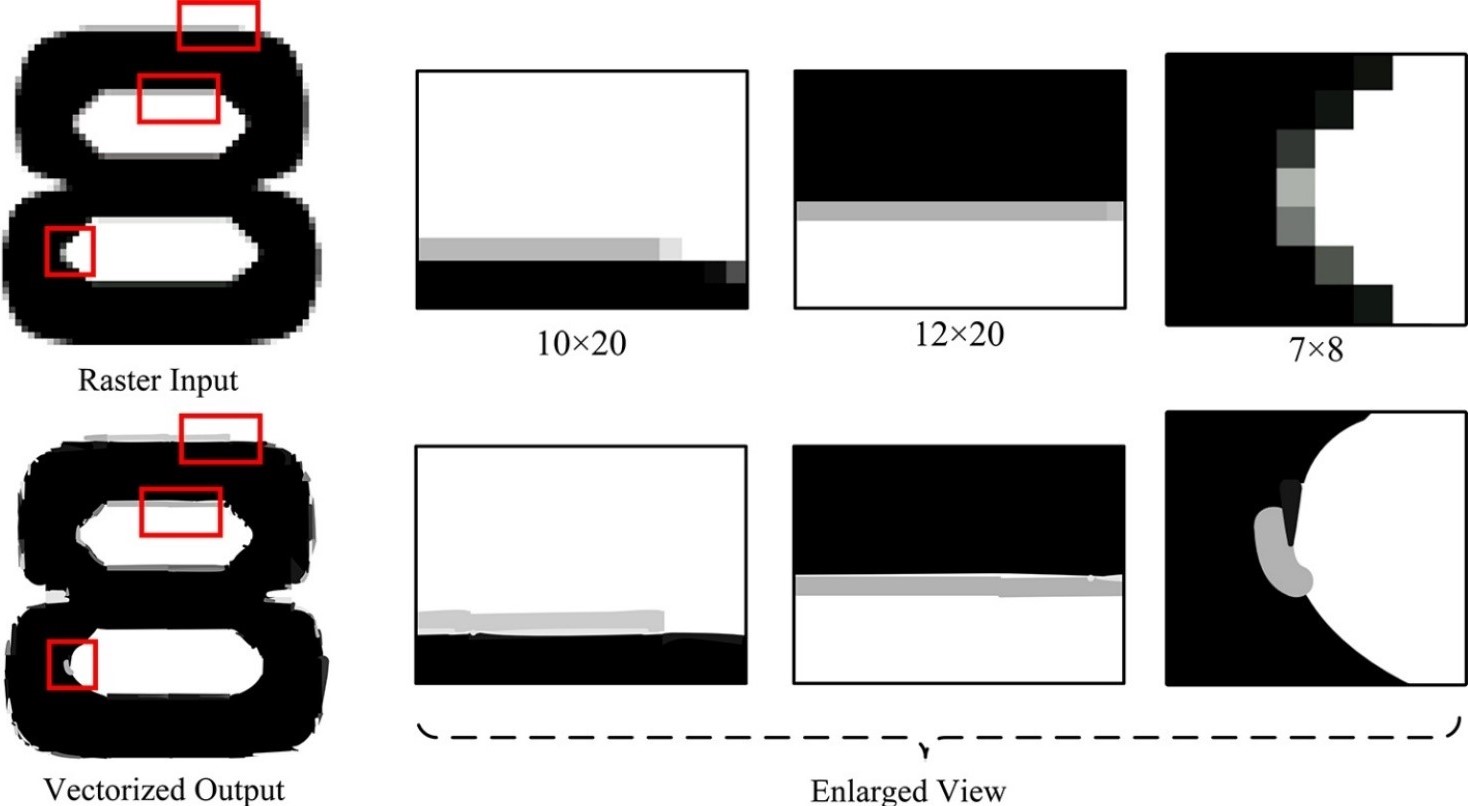}
\caption{The vectorized gray regions (red boxes) are generated intentionally, which aims to accurately follow the gray pixels (red boxes) in raster inputs. }
\label{fig:compare_s2}
\end{figure}

\begin{figure}[t]
\centering
\includegraphics[width=3.5 in]{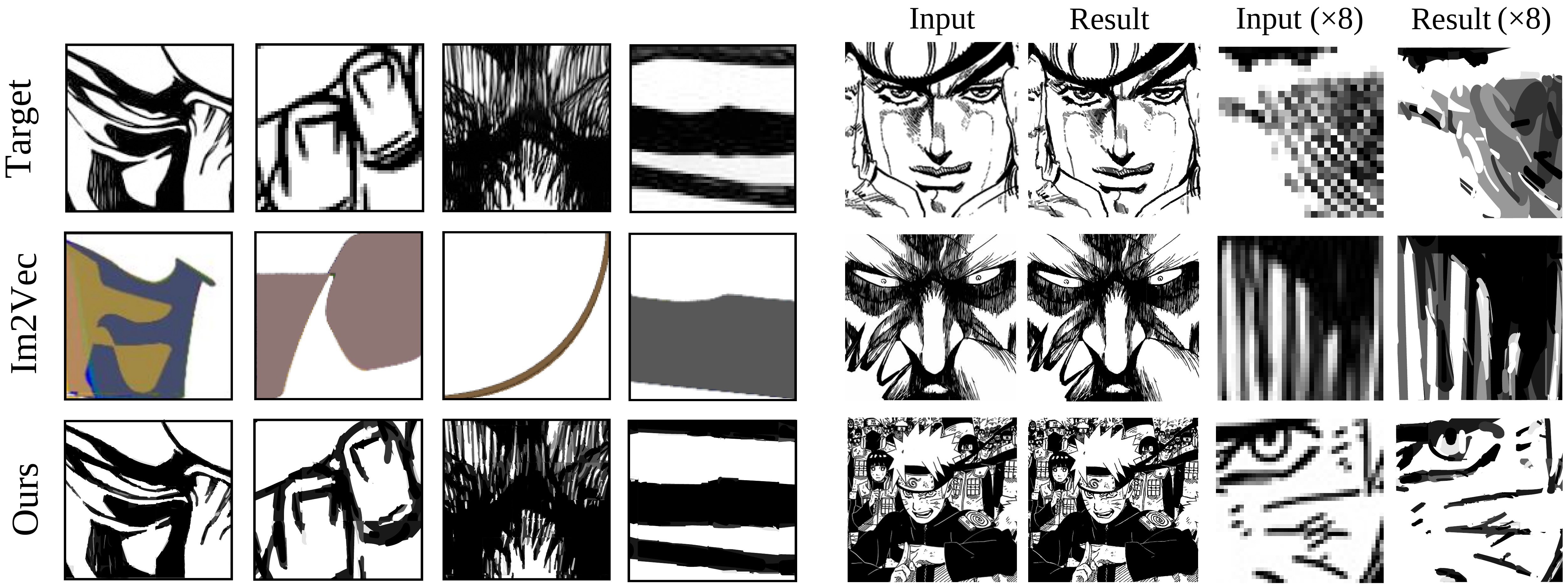}
\caption{\textbf{Left:} comparison with Im2Vec \cite{2021im2vec} that can be trained without vector supervised. \textbf{Right:} our vectorized results at different scales.}
\label{fig:compare_im2vec}
\end{figure}


 \renewcommand\arraystretch{1.3}
\begin{table*}[t]
\centering
\caption{Average vectorization accuracies for different manga types, including manga with sparse hatching (SH), dense hatching (DH), solid black strokes (SB), dense screentones (DS), and sparse screentones (SS). \textbf{Upper:} MSE. \textbf{Bottom:} SSIM. MMV \cite{TVCG2016manga} and Potrace \cite{potrace} excel at handling black-and-white (binarized) mangas. Especially, MMV is particularly adept at handling screentones. For binarized mangas, MMV, Portace, and Depixelizing \cite{depixelizing}  achieve the top-3 accuracies.
For gray-level mangas, Depixelizing and our approach generate the most accurate results in both visual perception and objective measurement. }
\small
\begin{tabular}{l|ccccc|ccccc}
\hline
\hline
\multirow{2}{*}{Method\textbackslash{}Manga type} & \multicolumn{5}{c|}{Binarized GT}                                                                          & \multicolumn{5}{c}{Gray-level GT}                                                                   \\ \cline{2-11} 
                                                  & \multicolumn{1}{c|}{SH} & \multicolumn{1}{c|}{DH} & \multicolumn{1}{c|}{SB} & \multicolumn{1}{c|}{DS} & SS & \multicolumn{1}{c|}{SH} & \multicolumn{1}{c|}{DH} & \multicolumn{1}{c|}{SB} & \multicolumn{1}{c|}{DS} & SS \\ 
\hline
Depixelizing                                      & \multicolumn{1}{l|}{0.0542}   & \multicolumn{1}{l|}{0.0568}   & \multicolumn{1}{l|}{0.0538}   & \multicolumn{1}{l|}{0.0554}   &  0.0588  & \multicolumn{1}{l|}{0.0702}   & \multicolumn{1}{l|}{0.0682}   & \multicolumn{1}{l|}{0.0713}   & \multicolumn{1}{l|}{0.0696}   & 0.0722   \\ 
\hline
Adobe                                             & \multicolumn{1}{l|}{0.1184}   & \multicolumn{1}{l|}{0.1365}   & \multicolumn{1}{l|}{0.1042}   & \multicolumn{1}{l|}{0.1285}   &  0.1162  & \multicolumn{1}{l|}{0.1367}   & \multicolumn{1}{l|}{0.1402}   & \multicolumn{1}{l|}{0.1338}   & \multicolumn{1}{l|}{0.1398}   &  0.1323   \\ 
\hline
Potrace                                           & \multicolumn{1}{l|}{ 0.0819 }   & \multicolumn{1}{l|}{ 0.0813 }   & \multicolumn{1}{l|}{0.0815}   & \multicolumn{1}{l|}{0.0843}   & 0.0825    & \multicolumn{1}{l|}{0.1145}   & \multicolumn{1}{l|}{0.1253}   & \multicolumn{1}{l|}{0.0933}   & \multicolumn{1}{l|}{0.1135}   & 0.1205    \\ 
\hline
MMV                                               & \multicolumn{1}{l|}{{0.0765}}   & \multicolumn{1}{l|}{ {0.0804}}   & \multicolumn{1}{l|}{0.0753}   & \multicolumn{1}{l|}{0.0795}   &  0.0778  & \multicolumn{1}{l|}{0.0883}   & \multicolumn{1}{l|}{0.0979}   & \multicolumn{1}{l|}{0.0865}   & \multicolumn{1}{l|}{0.0845}   &  0.0838  \\ 
\hline
\textbf{Ours}                                     & \multicolumn{1}{l|}{\textbf{0.0826}}   & \multicolumn{1}{l|}{ \textbf{0.0858 }}   & \multicolumn{1}{l|}{\textbf{0.0825}}   & \multicolumn{1}{l|}{\textbf{0.0848}}   &  \textbf{0.0834}  & \multicolumn{1}{l|}{\textbf{0.0812}}   & \multicolumn{1}{l|}{\textbf{0.0863}}   & \multicolumn{1}{l|}{\textbf{0.0794}}   & \multicolumn{1}{l|}{\textbf{0.0837}}   &  \textbf{0.0826}  \\ 
\hline
\hline
Depixelizing                                      & \multicolumn{1}{l|}{0.9554}   & \multicolumn{1}{l|}{0.9425}   & \multicolumn{1}{l|}{0.9607}   & \multicolumn{1}{l|}{0.9405}   &  0.9372   & \multicolumn{1}{l|}{0.9532}   & \multicolumn{1}{l|}{0.9573}   & \multicolumn{1}{l|}{0.9405}   & \multicolumn{1}{l|}{0.9513}   &  0.9472   \\ 
\hline
Adobe                                             & \multicolumn{1}{l|}{0.6203}   & \multicolumn{1}{l|}{0.5826}   & \multicolumn{1}{l|}{0.6312}   & \multicolumn{1}{l|}{0.5925}   &  0.6275  & \multicolumn{1}{l|}{0.6407}   & \multicolumn{1}{l|}{0.6152}   & \multicolumn{1}{l|}{0.6530}   & \multicolumn{1}{l|}{0.6223}   &  0.6512  \\ 
\hline
Potrace                                           & \multicolumn{1}{l|}{ 0.9327}   & \multicolumn{1}{l|}{ 0.8896 }   & \multicolumn{1}{l|}{0.9354}   & \multicolumn{1}{l|}{0.9180}   & 0.9212   & \multicolumn{1}{l|}{0.8241}   & \multicolumn{1}{l|}{0.7685}   & \multicolumn{1}{l|}{0.9075}   & \multicolumn{1}{l|}{0.8364}   &   0.8417 \\ 
\hline
MMV                                               & \multicolumn{1}{l|}{{0.9415}}   & \multicolumn{1}{l|}{ 0.8965}   & \multicolumn{1}{l|}{0.9442}   & \multicolumn{1}{l|}{0.9342}   &  0.9435  &
\multicolumn{1}{l|}{0.8532}   & \multicolumn{1}{l|}{0.7913}   & \multicolumn{1}{l|}{0.9205}   & \multicolumn{1}{l|}{0.8605}   & 0.8741    \\ 
\hline
\textbf{Ours}                                     & \multicolumn{1}{l|}{\textbf{0.9231}}   & \multicolumn{1}{l|}{ \textbf{0.8670} }   & \multicolumn{1}{l|}{\textbf{0.9252}}   & \multicolumn{1}{l|}{\textbf{0.8724}}   & \textbf{0.9193}   & \multicolumn{1}{l|}{\textbf{0.9352}}   & \multicolumn{1}{l|}{\textbf{0.8750}}   & \multicolumn{1}{l|}{\textbf{0.9261}}   & \multicolumn{1}{l|}{\textbf{0.8873}}   &  \textbf{0.9025}  \\ 
\hline
\hline
\end{tabular}
\label{tab:compare_algorithm-based_diffdata}
\end{table*}

\subsection{Comparison with algorithm-based methods}
We first compare our MARVEL with representative methods of algorithm-based vectorization and commercial packages, including Depixelizing \cite{depixelizing}, Adobe Illustrator \cite{adobe} (high fidelity mode), Potrace \cite{potrace}, and MMV (Mesh-based Manga Vectorization) \cite{TVCG2016manga}.
We randomly select 50 raster inputs in Deepmanga, and generate 50 vectorized outputs by each compared method.

\textbf{Visual perception:} the vectorization results of each method are shown in Figure \ref{fig:compare-algorithm-based} and Figure \ref{fig:compare-algorithm-based-large}. 
The outputs of Adobe Illustrator \cite{adobe} typically lose contents (Figure \ref{fig:compare-algorithm-based} red boxes) of inputs. Potrace \cite{potrace} neither handles multiple colors nor preserves accurate pixel details. MMV \cite{TVCG2016manga} excels at vectorizing screentones, yet, for complex input manga without screentones, MMV typically produces binary colors and ignores gray regions, which reduces the vectorization accuracy. In Section IV-D, we detailly summarize the advantages and disadvantages of MMV. 

In addition, the algorithm-based methods need to design tailored algorithms for different scenarios, which typically include complex image segmentation and curve-fitting. Our method aims to train an end-to-end neural model that can replace the algorithm design, image segmentation, and curve fitting.

\textbf{Accuracies of vectorizing different manga types:} for a fair comparison, first, we divide our test data into five manga categories, including manga with sparse hatching (SH), dense hatching (DH), solid black strokes (SB), dense screentones (DS), and sparse screentones (SS). Next, each ground truth (GT) is processed into two types, that is, binarized GT $G^b$ and multi-gray levels GT $G^g$. Then, we vectorize $G^b$ and $G^g$ by each algorithm-based method, respectively. 

The average vectorization accuracies are summarized in Table \ref{tab:compare_algorithm-based_diffdata}. We observe that MMV \cite{TVCG2016manga} and Potrace \cite{potrace} excel at handling black-and-white (binarized) mangas. Especially, MMV is particularly adept at handling screentones. For binarized mangas, MMV, Portace, and Depixelizing \cite{depixelizing}  achieve the top-3 accuracies.
For multi-gray level manga, Depixelizing \cite{depixelizing} and our approach generate the most accurate results in both visual perception and objective measurement. Yet, Depixelizing converts pixels to square vector grids and only smooths several critical paths, which incurs the vectorization results to look as blurry as the raster inputs when zooming in (red boxes in Figure \ref{fig:compare-algorithm-based-large}). By contrast, our MARVEL achieves a better balance between accuracy and smoothness.

\renewcommand\arraystretch{1.2}
\begin{table}[t]
\vspace{0.5cm}
\centering
\caption{Average vectorization accuracies compared with learning-based methods.}
\small
\begin{tabular}{ m{1.8cm} |m{1cm} m{1.3cm} | m{0.9cm} m{1.3cm} }
\hline
\hline
Index            & \multicolumn{2}{c|}{MSE} & \multicolumn{2}{c}{SSIM} \\ \hline
\footnotesize{Method}\textbackslash{}\footnotesize{Dataset} & F\scriptsize{ONTS}      & $\!\!\!\!$DeepManga  & F\scriptsize{ONTS}      & $\!\!\!\!$DeepManga    \\
\hline
SVG-VAE                       & $\!$0.3815   &    \normalsize{$ \ \ \ \! \bm{\times}$}        & 0.4328    &   \normalsize{$\ \ \ \! \bm{\times}$}     \\
DeepSVG                       & $\!$0.3433    &    \normalsize{$\ \ \ \! \bm{\times}$}       & 0.5238     &   \normalsize{$\ \ \ \! \bm{\times}$}           \\
Im2Vec                        & $\!$0.0532    & 0.2445    & 0.8083     & 0.3879      \\
LIVE                        & $\!$0.0293    & 0.1672    & 0.8852     & 0.6787      \\
\textbf{Ours}                          & $\!$\textbf{0.0261}     & \textbf{0.0825}  & \textbf{0.9303}     & \textbf{0.8968}       \\
\hline
\hline
\end{tabular}
\label{tab:compare_font}
\end{table}

\begin{figure*}[t]
\centering
\includegraphics[width=7.2 in]{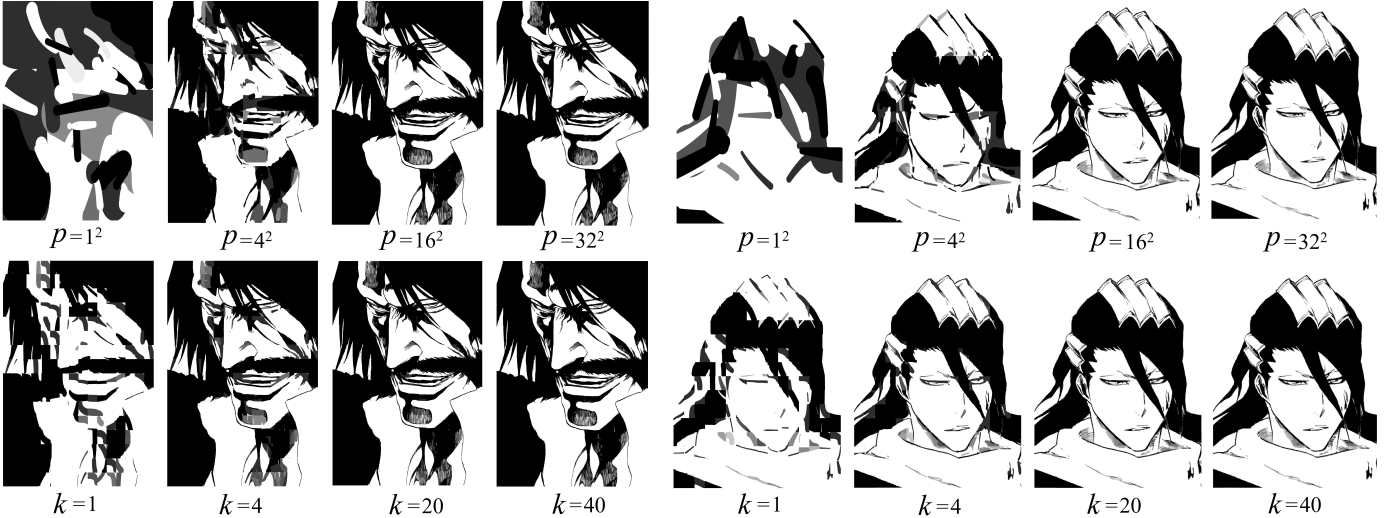}
\caption{Our vectorized results under different settings of $p$ and $k$. \textbf{Upper:} influence of $p$, when $k=40$. \textbf{Bottom:} influence of $k$, when $p=16^2$. Our vectorization accuracies are increasing with $p$ and $k$ through subjective visual perception, and the objective evaluation is shown in Figure \ref{fig:sp2}.}
\label{fig:sp}
\end{figure*}

\begin{figure}[t]
\centering
\includegraphics[width=3.5 in]{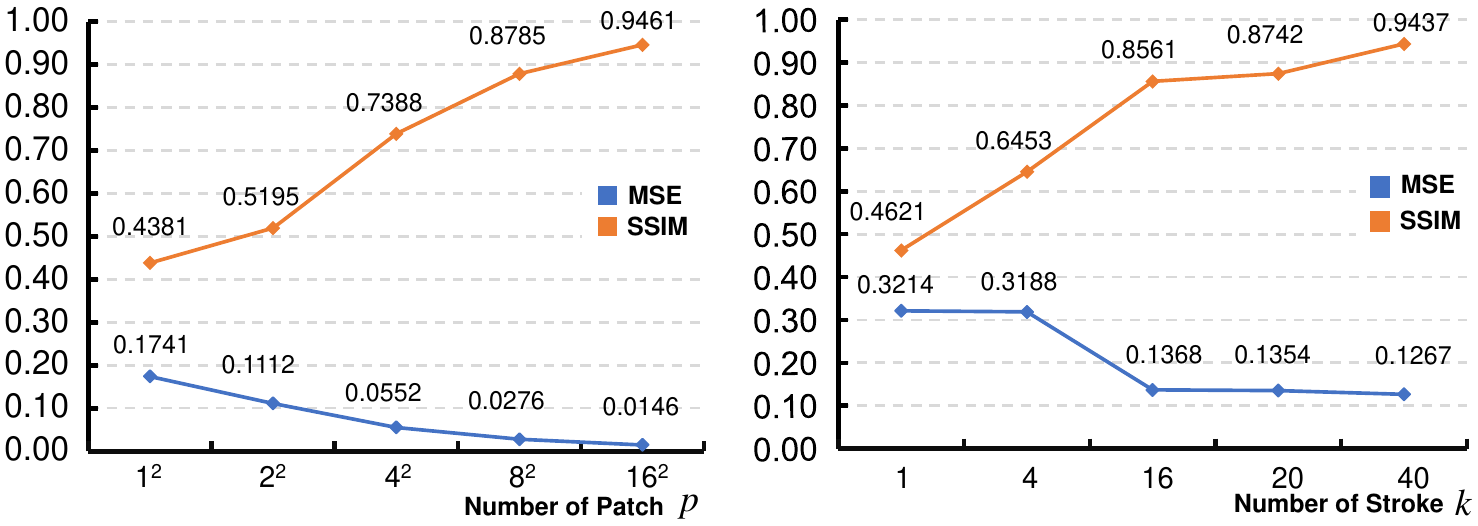}
\caption{Average vectorization accuracies in 100 random samples, under different settings of $p$ and $k$. The accuracies are proportional to $p$ and $k$, and the growth rates of accuracies are decreased when $k$ is higher than a threshold (i.e., $k$$\geqslant$$16$)}
\label{fig:sp2}
\end{figure}

\subsection{Comparison with learning-based methods}
Next, we compare our MARVEL with representative learning-based vectorization methods, including SVG-VAE \cite{svgvae}, DeepSVG \cite{deepsvg}, Im2Vec \cite{2021im2vec}, and LIVE \cite{ma2022towards} (N=256).

\textbf{Vectorization on simple structures:} we compare the vectorization accuracies on datasets with simple image structures (e.g., F\scriptsize{ONTS}\normalsize{}).
Figure \ref{fig:compare_s} and Table \ref{tab:compare_font} show the comparison results in visual perception and objective indexes respectively.

Compared with other learning-based methods, our vectorization results have high accuracies in objective and subjective, and reaches the state-of-the-art level for vectorizing images with simple structures. Moreover, as shown in Figure \ref{fig:compare_s2}, the artifacts-like vectorized gray regions (red boxes) are generated intentionally, which aims to accurately follow the gray pixels (red boxes) in raster inputs. This phenomenon is caused by unclear pixel outlines in raster inputs, and then our method will consider these unclear pixels as the vectorization targets, when setting a large $p$ to increase the detail level.

\textbf{Vectorization on complex structures:}~utilizing 50 random cropped inputs of sizes $128$$\times$$128$ in Deepmanga, we compare the performance of vectorizing complex structures. We only compare our method with Im2Vec \cite{2021im2vec} (trained by dataset DeepManga) and LIVE \cite{ma2022towards}, since DeepSVG \cite{deepsvg} and SVG-VAE \cite{svgvae} cannot be trained without the vector supervision.~Moreover, Reddy et al. \cite{2021im2vec} have proved that Im2Vec outperforms DeepSVG and SVG-VAE on vectorization accuracies.

Comparison results are shown in Figure \ref{fig:compare-algorithm-based-large}, Figure \ref{fig:compare_im2vec} and Table \ref{tab:compare_font}. Im2Vec cannot handle target with complex content, and LIVE (N=256) may miss the contents of inputs and produces gray outputs from white targets (Figure \ref{fig:compare-algorithm-based} red boxes). In addition, LIVE is an iteration-based method that optimizes each vector element using 500 iterations (1-5 seconds per iteration, increasing with the indexes of vector elements), which takes an average of 25.88 hours to vectorize one image, when setting N=256 and N indicates the maximum number of vector elements.


\renewcommand\arraystretch{1.6}
\begin{table}[t]\footnotesize 
\centering
\caption{Differences between our approach and state-of-the-art learning-based vectorization methods (i.e., SVG-VAE \cite{svgvae}, DeepSVG \cite{deepsvg}, Im2Vec \cite{2021im2vec}, and LIVE \cite{ma2022towards}).}
\begin{tabular}{m{2.7cm}<{\raggedright}|m{0.8cm}<{\centering}|m{0.7cm}<{\centering}|m{0.7cm}<{\centering}|m{0.6cm}<{\centering}|m{0.6cm}<{\centering}}
\hline
\hline
\textbf{$\!\!\!\!$Performance}\textbackslash{}\textbf{Method} &$\!\!\!$\scriptsize{SVG-VAE}$\!\!\!$&$\!\!\!$\scriptsize{DeepSVG}$\!\!\!$&$\!\!\!$\scriptsize{Im2Vec}$\!\!\!$&$\!\!\!$\scriptsize{LIVE}$\!\!\!$&$\!\!\!$\scriptsize{\textbf{Ours}}$\!\!\!$  \\
\hline

$\!\!\!\!$Vectorize simple texture  & \normalsize\checkmark & \normalsize\checkmark & \normalsize\checkmark & \normalsize\checkmark  & \normalsize\checkmark  \\
\hline
$\!\!\!\!$Vectorize complex texture$\!\!$& \scriptsize{\XSolidBrush} & \scriptsize{\XSolidBrush} & \scriptsize{\XSolidBrush} & \normalsize\checkmark & \normalsize\checkmark  \\
\hline
$\!\!\!\!$W/O vector supervision     & \scriptsize{\XSolidBrush} & \scriptsize{\XSolidBrush} & \normalsize\checkmark & \normalsize\checkmark & \normalsize\checkmark  \\
\hline
$\!\!\!\!$Average time cost& 1.12s & 0.05s & 14.37s &$\!\!$25.88h  &$\!$48.23s  \\
\hline
\hline
\end{tabular}
\label{tab:compare}
\end{table}
\renewcommand\arraystretch{1}

To summarize, as shown in Table \ref{tab:compare}, first, our method outperforms compared learning-based methods in the accuracies of vectorizing simple and complex structures. Second, our method has the merit that only needs supervision from raster images without high-cost vector annotation. Third, compared with LIVE \cite{ma2022towards} which can vectorize complex contents (average 25.88 hours when N=256), our method takes much less time (average 48.23 seconds when $p$$=$$16^2$, $k$$=$$20$)$^1$. \footnote{$^1$Detailed evaluation of our time-cost is shown in Section \ref{sec:timecost}.}

\subsection{Comparison with manga and line-art vectorization methods}
\begin{figure}[t]
\centering
\includegraphics[width=3.5 in]{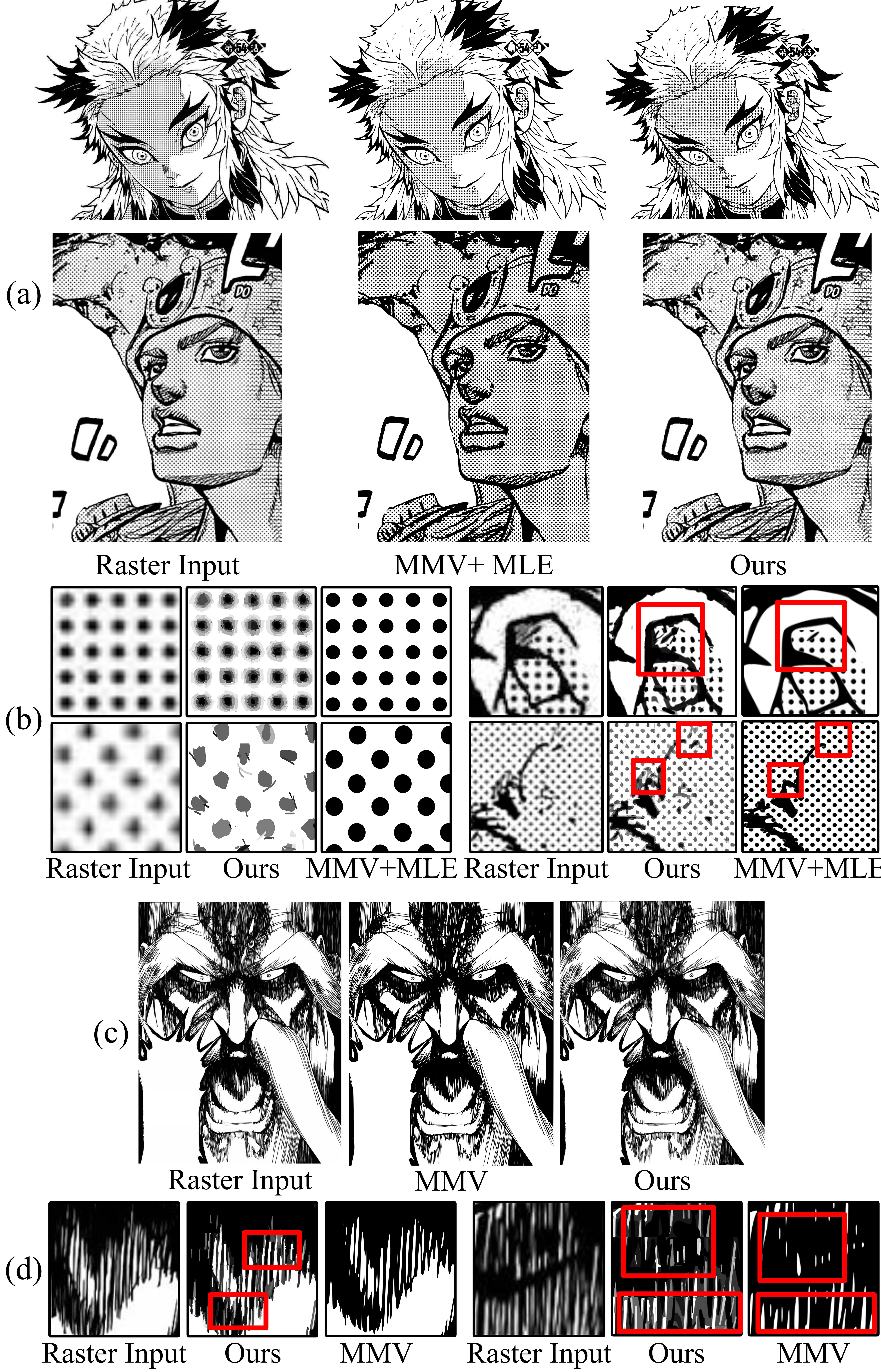}
\caption{Comparison with MMV \cite{TVCG2016manga}. (a) Vectorization of manga with screentones \cite{TOG2021r3}. (b) Enlarged view of (a). (c) Vectorization of manga with complex structures (e.g., dense hand-drawing shadows. (d) enlarged view of (b). }
\label{fig:compareMMV}
\end{figure}

\textbf{Manga vectorization:} as shown in Figure \ref{fig:compareMMV}, we first compare and discuss our performance with manga vectorization method MMV \cite{TVCG2016manga}. In implementing MMV, we adopt MLE (Manga Line Extraction \cite{tog2017r1}) instead of the original screentones detection method.
MMV offers some benefits, including the ability to extract key vector paths for easy shape editing and smaller file sizes (averaging 126.35KB), and MMV excels at handling screentones [Figure \ref{fig:compareMMV}(a)(b)].
However, it has some limitations. First, as shown in Figure \ref{fig:compareMMV}(d), MMV sacrifices performance in preserving similarities (Table \ref{tab:compare_algorithm-based_diffdata}) and color levels (i.e., only preserves binary colors except for screentones while ignoring gray regions). Second, it struggles to preserve complex structures (e.g., dense hand-drawing shadow lines [Figure \ref{fig:compareMMV}(d) red boxes]), which may lose some contents and stroke lines [Figure \ref{fig:compareMMV}(b) red boxes].
By contrast, our advantages are as follows. First, we preserve high similarities and color levels with input mangas. Second, our method excels at handling complex manga structures. Nevertheless, our method has two limitations. First, it cannot extract key paths, resulting in larger file sizes (averaging 734.54KB). Second, Second, it lacks a dedicated setting to process manage screentones.

\begin{figure}[t]
\centering
\includegraphics[width=3.2 in]{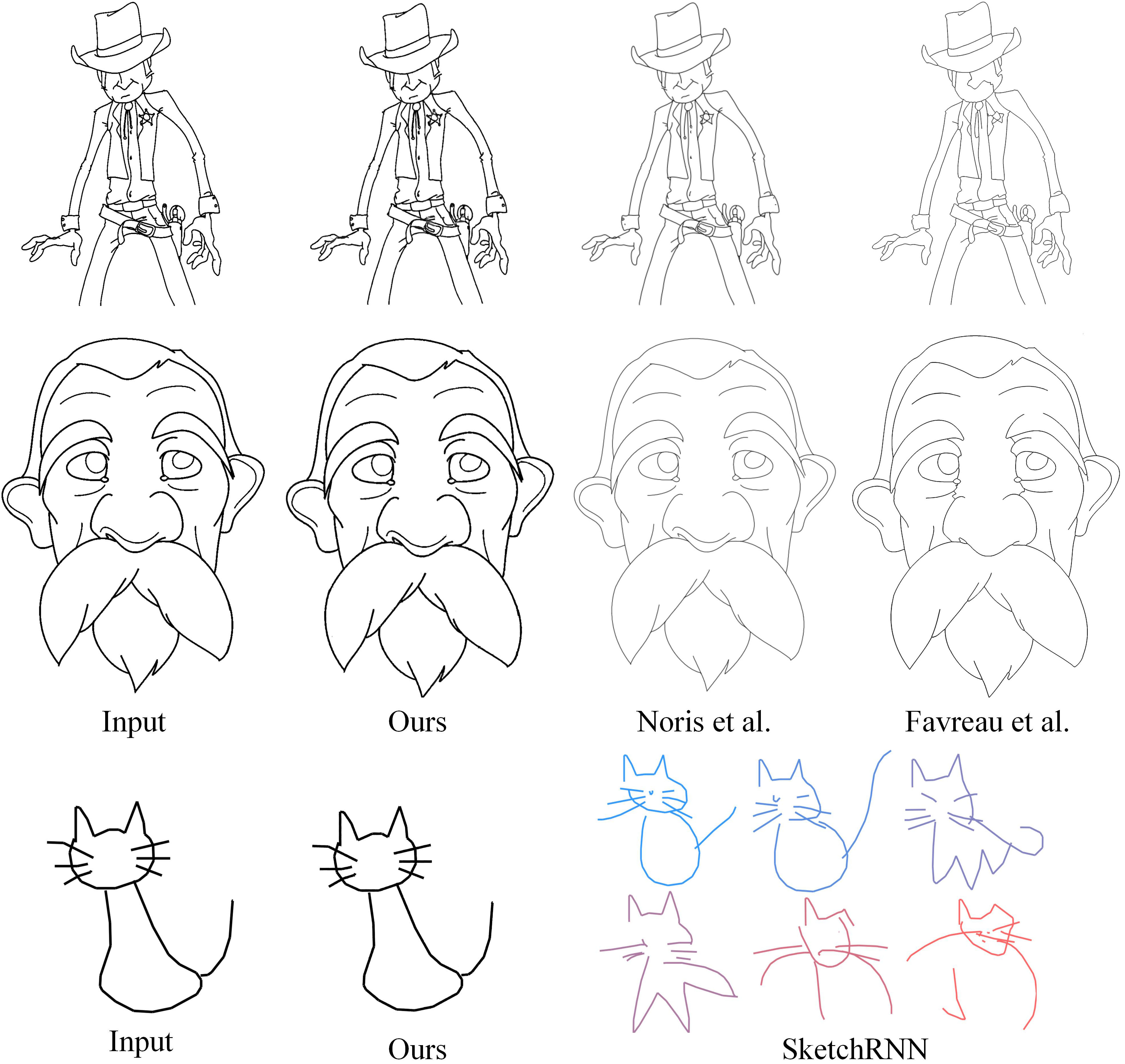}
\caption{Comparison with representative line-art vectorization methods, including Noris et al. \cite{tog2013cartoon}, Favreau et al. \cite{favreau2016fidelity}, and SketchRNN \cite{ha2017neural}.}
\label{fig:compareline}
\end{figure}

\begin{figure}[t]
\centering
\includegraphics[width=3.5 in]{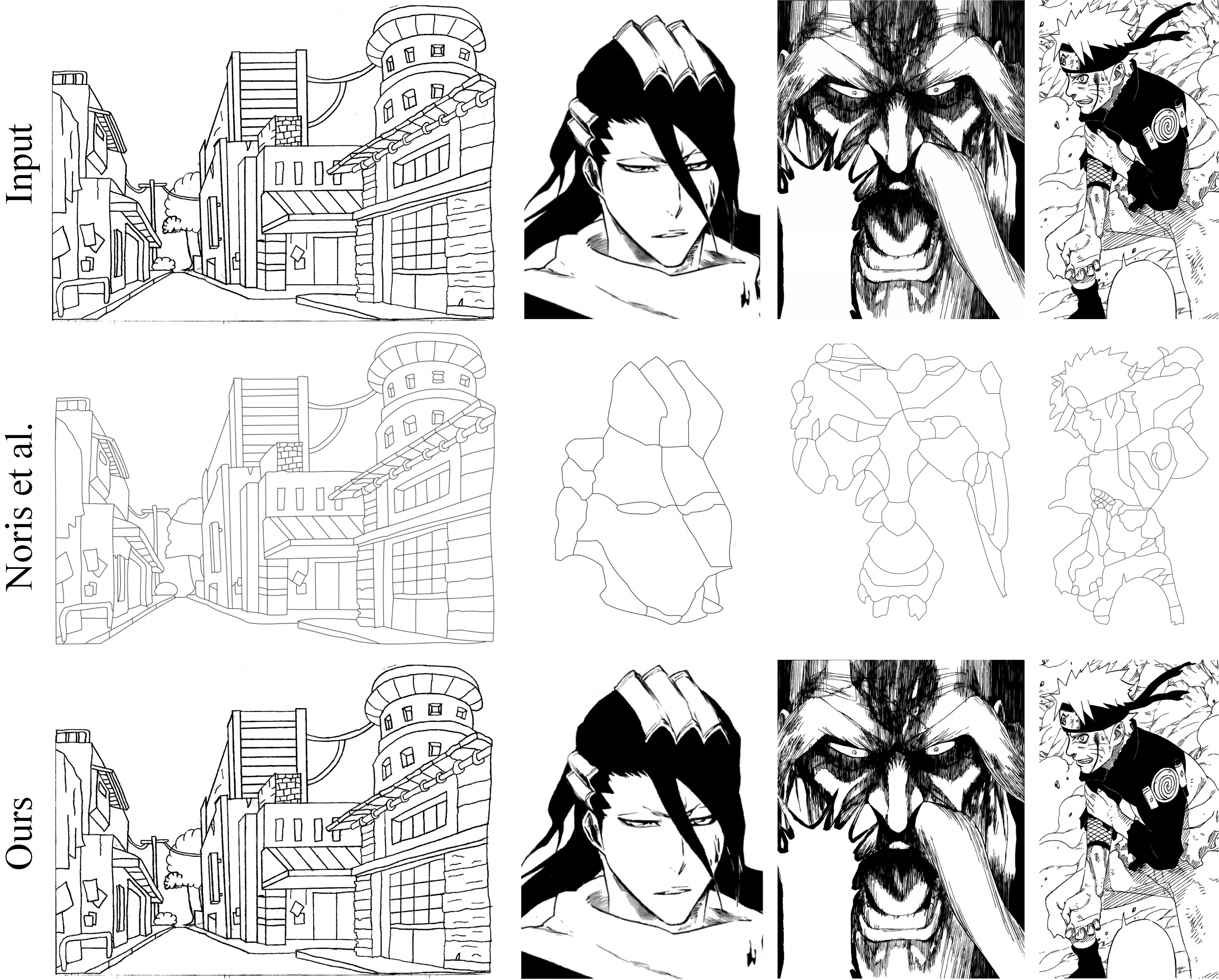}
\caption{Comparison with Noris et al. \cite{tog2013cartoon} for vectorizing line-arts and complex mangas.}
\label{fig:compareline2}
\end{figure}

\textbf{Line-art vectorization:} we next compare and discuss our method with representative line-art vectorization methods, including Noris et al. \cite{tog2013cartoon}, Favreau et al. \cite{favreau2016fidelity}, and SketchRNN \cite{ha2017neural}. As shown in Figure \ref{fig:compareline} and Figure \ref{fig:compareline2}, these methods are typically employed to extract the vector centerline of the line-art drawings (e.g., sketches, clean line drawings). SketchRNN \cite{ha2017neural} reconstructs vector paths according to the input sketches, and outputs a series of vectorized results with similar structures and paths.
By contrast, our method aims to preserve high similarities with inputs and cannot simplify contents to key paths.

\begin{figure}[t]
\centering
\includegraphics[width=2.8 in]{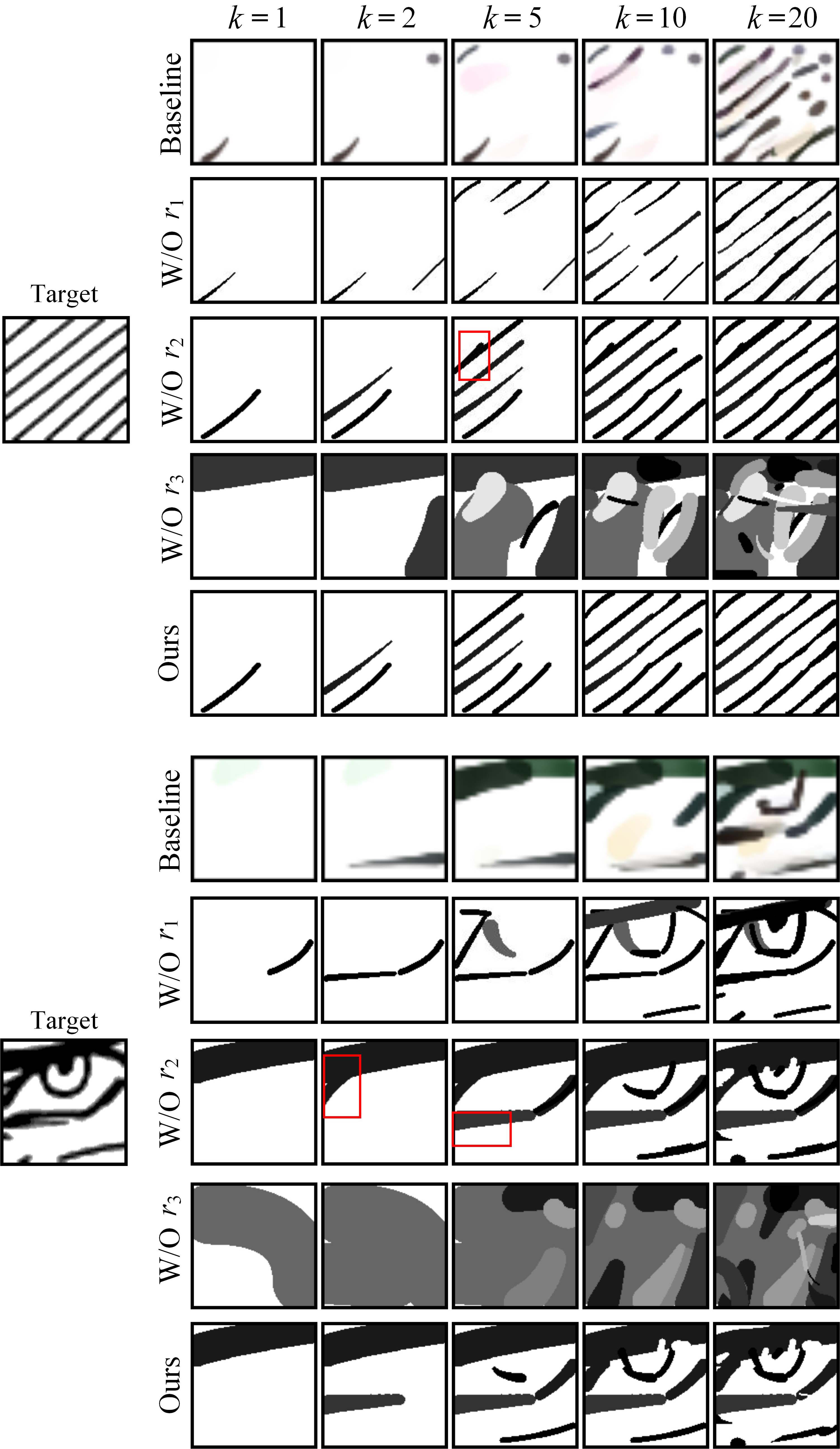}
\caption{Ablation study of rewards settings, including $\mathcal{R}_{L2}$ reward (baseline \cite{l2p}), and results without $r_1$, $r_2$, and $r_3$ respectively. Objective comparison is shown in Figure \ref{fig:ablation_object}.}
\label{fig:ablation}
\end{figure}

\begin{figure}[t]
\centering
\includegraphics[width=3.5 in]{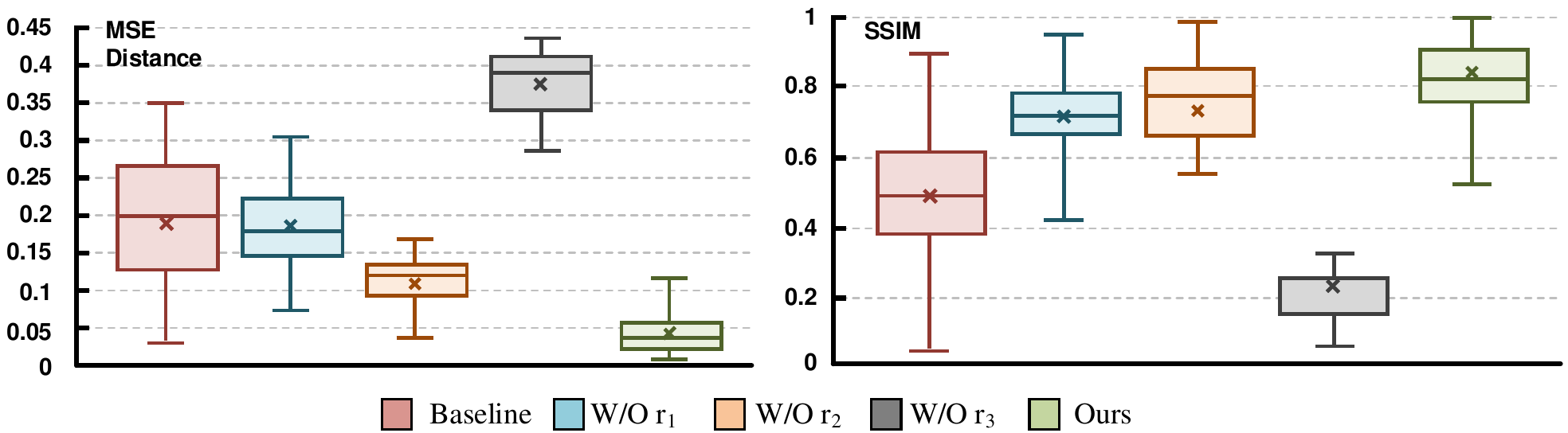}
\caption{Object similarities between inputs and outputs in 100 random samples when ${k}$$=$$20$ and $p$$=$$1$$\times$$1$.}
\label{fig:ablation_object}
\end{figure}



\subsection{Ablation study}
\textbf{Influence of patches and strokes:} in MARVEL, there are two manners to increase the vectorization accuracy while compromising the time cost. First, dividing a target image into small patches and vectorizing each of them. Second, increasing the maximum number of stroke lines. Accordingly, we conduct an experiment to evaluate the influence of the number of patches $p$ and the maximum number of stroke lines $k$.


In Figure \ref{fig:sp}, we display our vectorized results under different settings of $p$ and $k$, and the average vectorization accuracies in 100 random samples are shown in Figure \ref{fig:sp2}. When setting a small $p$, the dense textures may be vectorized as a rough gray region to decrease the file size. When setting a large $p$, the textures or strokes in vectorized results are obvious and clean.
Our vectorization accuracies are proportional to $p$ and $k$, and the growth rates of accuracies are decreased when $k$ is higher than a threshold (i.e., $k$$\geqslant$$16$). Empirically, setting $p$$=$$16$$\times$$16$ and $k$$=$$20$ is enough to produce results with high accuracies.

\textbf{Influence of rewards setting:}  to evaluate our rewards' effectiveness in improving accuracies, we train a DRL agent
 leveraging each of $\mathcal{R}_{L2}$ (baseline \cite{l2p}), $(r_2+r_3)$, $(r_1+r_3)$, $(r_1+r_2)$, and $(r_1+r_2+r_3)$ respectively. As shown in Figure \ref{fig:ablation}, the visual comparison demonstrates that without $r_1$, the agent tends to predict small and inaccuracy strokes. Without $r_2$, the differences between the current and next canvas are decreased, which incurs the overlap of strokes (marked with red boxes). Without $r_3$, the combination of strokes cannot progressively be similar to the ground truth.
The scatter and box charts in Figure \ref{fig:ablation_object} display the measured similarities between inputs and outputs in 100 random samples when ${k}$$=$$40$ and $p$$=$$1$$\times$$1$, which also shows our proposed rewards can effectively improve the accuracy of stroke lines.
 To sum up, each term of our proposed rewards is effective to predict more accurate strokes in each timestep, and significantly outperforms the baseline.

\begin{figure}[t]
\vspace{0.5cm}
\centering
\includegraphics[width=3.5 in]{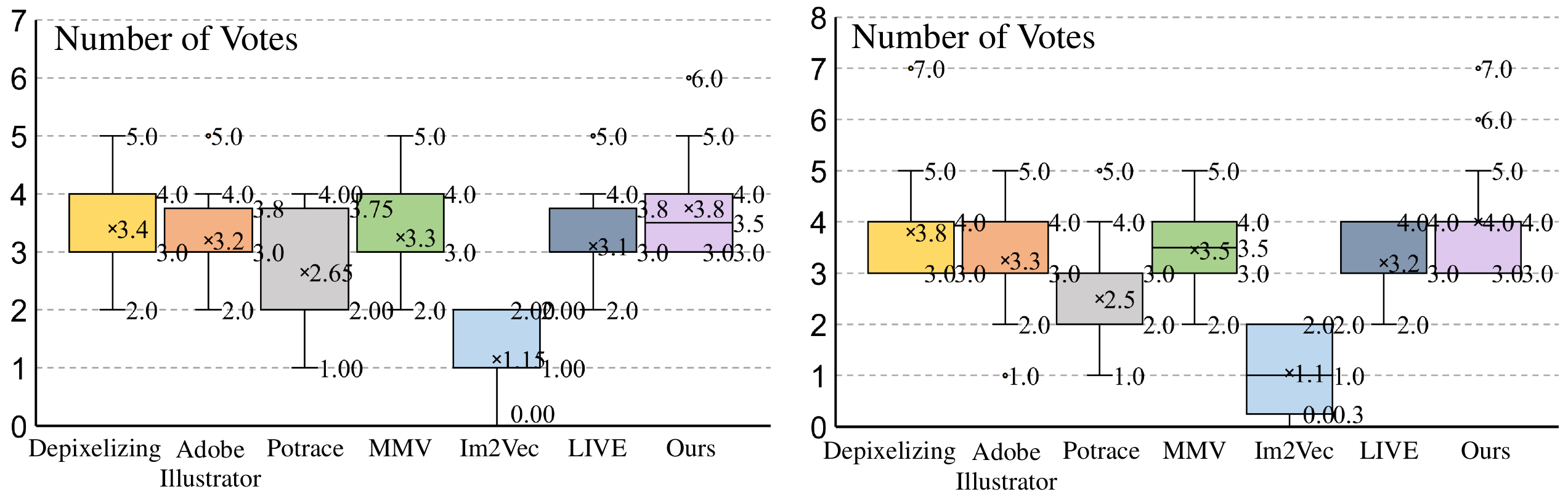}
\caption{Comparison with related methods in manga vectorization. \textbf{Left:} voting results for visual effect. \textbf{Right:} voting results for vectorization accuracy. }
\label{fig:userstudy1}
\end{figure}

\begin{figure}[t]
\centering
\includegraphics[width=3.5 in]{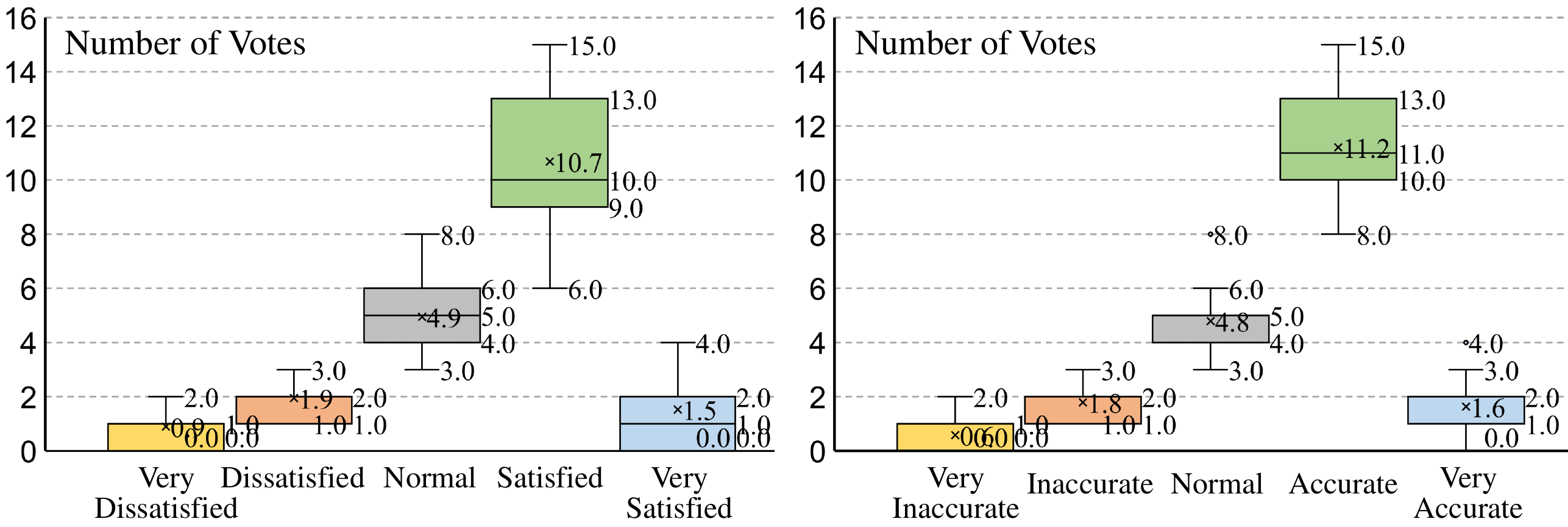}
\caption{User studies on visual effect and vectorization accuracy of our results. \textbf{Left:} voting results for visual effect. \textbf{Right:} voting results for vectorization accuracy.}
\label{fig:userstudy}
\end{figure}

\renewcommand\arraystretch{1.2}
\begin{table}[t]
\centering
\small
\caption{Paired two-tailed t-test of our user studies. \textbf{Upper:} voting results for visual effect. \textbf{Bottom: } voting
results for vectorization accuracy.  }
\small
\begin{tabular}{l l c c c}
\hline
\hline
\textbf{Method}            & $\mathbf{\bar{X}\pm S}$ & D & $t$ &  $p$ \\
\hline
Depixelizing      & $3.40 \pm 0.68$  & $-0.35$  & $-1.277 $                & $0.217 $                 \\
Adobe Illustrator & $3.20 \pm 0.70$  & $-0.55$  & $-1.868 $                & $0.077$                  \\
Potrace           & $2.65 \pm 0.93$  & $-1.10$  & $-3.240$                 & $0.004 $                 \\
MMV               & $3.25 \pm 0.72$  & $-0.50$  & $-2.032$                 & $0.056 $                 \\
Im2Vec            & $1.15 \pm 0.75$  & $-2.60$   & $-9.133 $               & $<$$0.001$       \\
LIVE              & $3.10 \pm 0.79$  & $-0.65$  & $-2.668$                & $0.015 $                 \\
Ours              & $3.75 \pm 0.91$  & $0.00$  &                        &   \\
\hline
\hline
Depixelizing      & $3.80 \pm 1.01 $                        & $-0.20$                  & $-0.535$                & $0.599$                 \\
Adobe Illustrator & $3.25 \pm 1.02 $                        & $-0.75$                 & $-1.861$                & $0.078$                 \\
Potrace           & $2.50 \pm 0.89 $                        & $-1.50$                  & $-4.177$                & $0.001$                 \\
MMV               & $3.45 \pm 0.76  $                       & $-0.55$                  & $-1.764$                & $0.094$                     \\
Im2Vec            & $1.05 \pm 0.76 $                        & $-2.95$                 & $-9.460$                 & $<$$0.001$               \\
LIVE              & $3.20 \pm 0.62 $                        & $-0.80$                  & $-3.559$                & $0.002$                \\
Ours              & $4.00 \pm 1.08$                         & $0.00$ &      &   \\
\hline
\hline
\end{tabular}
\label{tab:t-test-visual}
\end{table}

\subsection{User study:}
Below we conduct two user studies to subjectively evaluate our results' visual effects and accuracies, and 20 volunteers are invited to participate in the studies.

\textbf{Comparison with related methods:} after observing 20 samples produced by each compared method, all volunteers are asked to vote for the two methods whose results are most visual-pleasant and most accurate respectively. The study results in Figure \ref{fig:userstudy1} indicate that on average, 3.75 (18.75$\%$) and 4.00 (20.00$\%$) volunteers believe our results have the highest visual quality and accuracy respectively.
Table \ref{tab:t-test-visual} shows the paired two-tailed t-test of our user studies in Figure \ref{fig:userstudy1}, where $\mathbf{\bar{X}}$, $\mathbf{S}$, and $\mathbf{D}$ indicate mean, standard deviation, and difference of mean, respectively.
To sum up, compared with algorithm-based and learning-based methods, our approach achieves state-of-the-art performance on visual effect and vectorization accuracy.

\textbf{Subjective perception of visual effect and accuracy:} another user study is to evaluate the subjective perception of our approach in visual effect and accuracy. First, we randomly generate 20 paired samples containing raster inputs and corresponding vectorized outputs. Then, the samples are displayed on a webpage, and 20 volunteers are invited to finish two tasks after observation. The first task is to vote for the satisfaction levels of visual effect, and another task is to vote for the accuracy levels.

As shown in Figure \ref{fig:userstudy}, for visual effect, 10.68$\,$/$\,$4.97 ($53.42\% \,$/$ \,24.85\%$) volunteers on average have voted for the satisfied/normal levels. For accuracy, 11.21$\,$/$\,$4.79 ($56.05\%\,$/$\,23.95\%$) volunteers on average have voted for the accuracy/normal levels.
Accordingly, the study results show that our vectorized results satisfy the requirements of most users in visual effect and accuracy.

\begin{figure}[t]
\vspace{0.5cm}
\centering
\includegraphics[width=3.5 in]{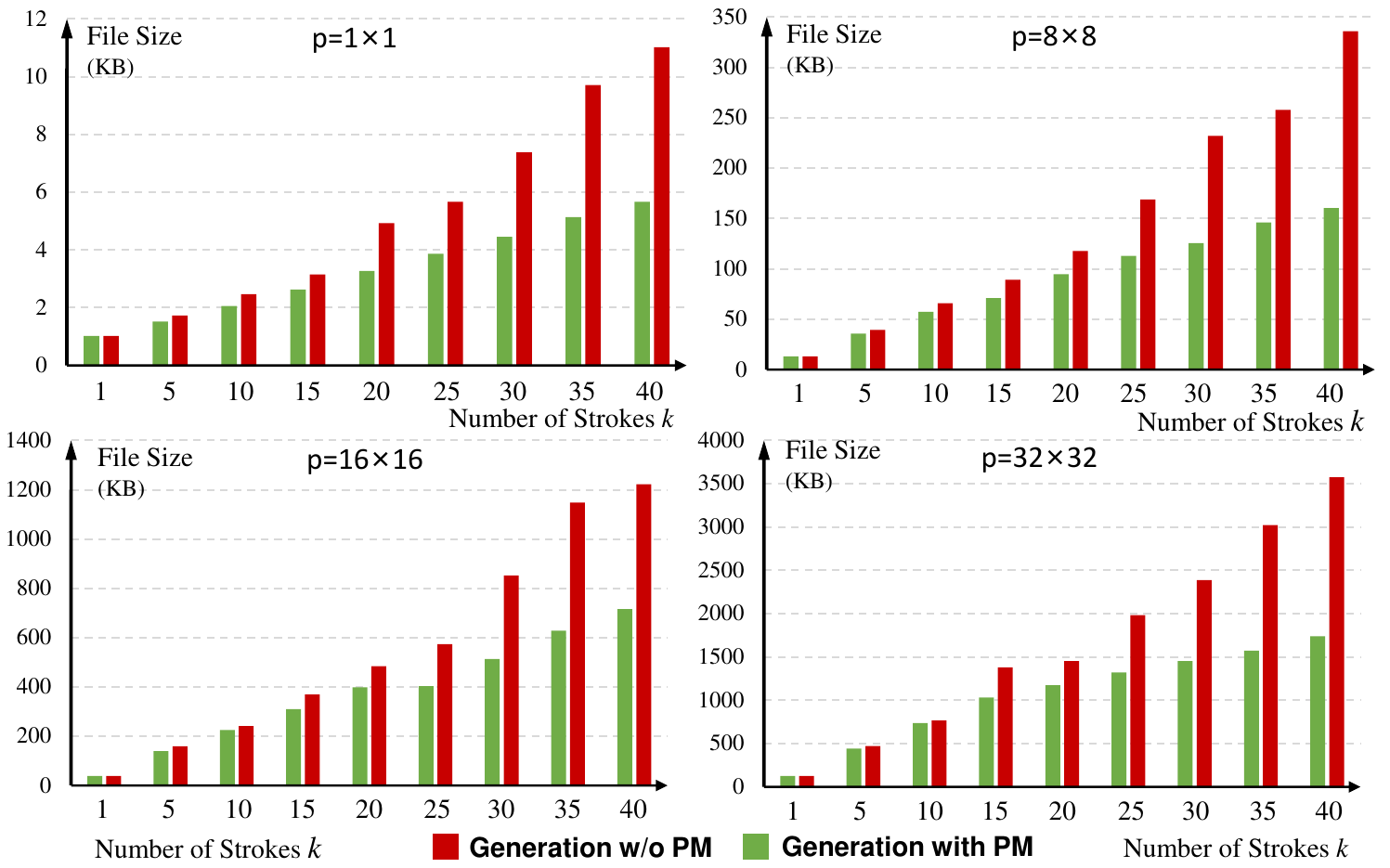}
\caption{Influences of the pruning module (PM) on file sizes, under different settings of $p$ and $k$. The use of PM significantly reduces file sizes, and the reduced file sizes are proportional to $k$ since the produced redundant and erroneous strokes are increasing with $k$.}
\label{fig:filesize}
\end{figure}

\begin{figure}[t]
\vspace{0.2cm}
\centering
\includegraphics[width=3.5 in]{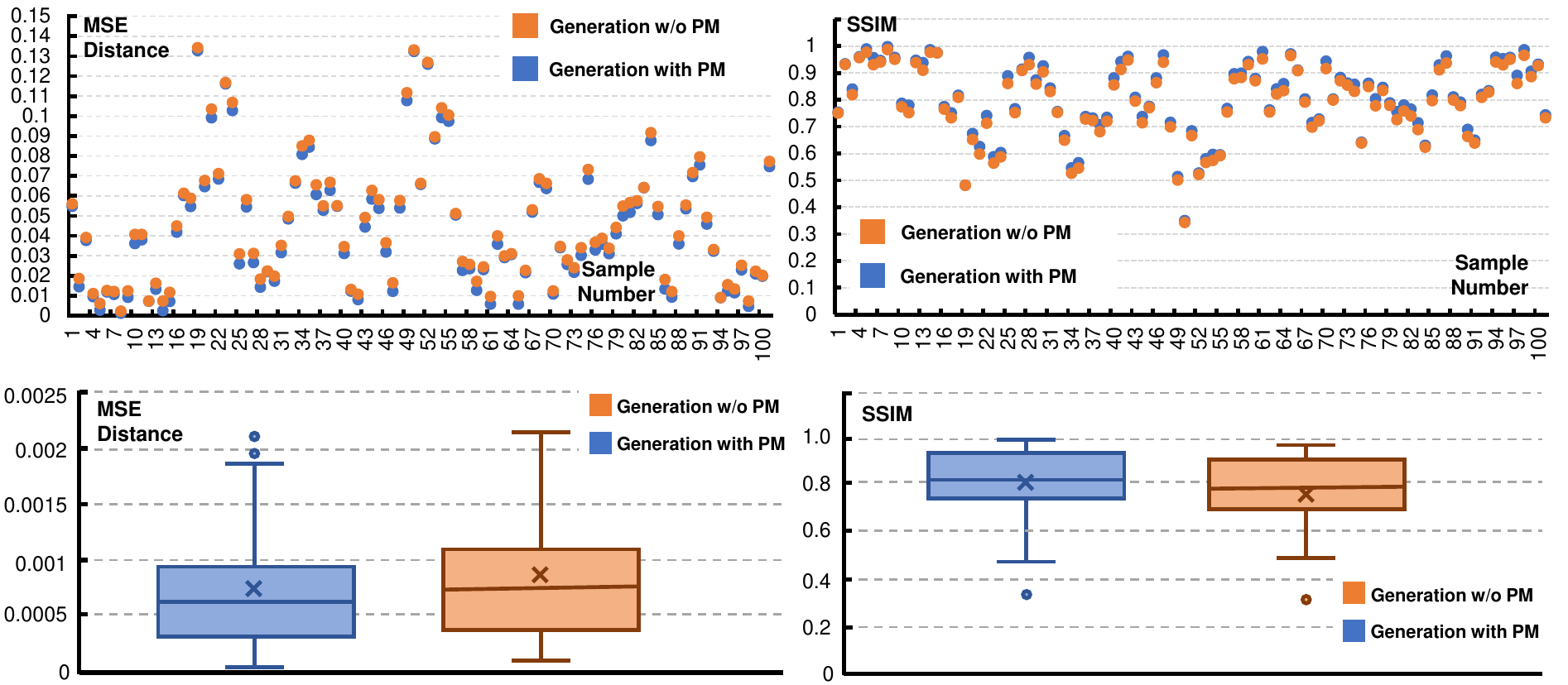}
\caption{Objective evaluation of average similarity of 100 paired samples (generation with PM and without PM), when $p=32^2$ and $k=40$. The results demonstrate that using PM does not compromise the visual similarity.}  
\label{fig:visualsimilarity}
\end{figure}

\renewcommand\arraystretch{1.5}
\begin{table}[t]
\centering
\vspace{0.6cm}
\caption{Time costs (in seconds) under different settings of $p$ and $k$. \textbf{Upper}: time costs of vectorization w/o PM. \textbf{Bottom}: time costs of PM processing.}
\scriptsize
\begin{tabular}{m{0.2cm}<{\centering}|m{0.3cm}<{\centering} m{0.5cm}<{\centering} m{0.5cm}<{\centering}m{0.5cm}<{\centering} m{0.5cm}<{\centering} m{0.5cm}<{\centering}m{0.5cm}<{\centering}m{0.5cm}<{\centering}m{0.5cm}<{\centering}}
\hline
\hline
\!\!\!$p$\verb|\|$k$  & $\mathbf{1}$    & $\mathbf{5}$     & $\mathbf{10}$    & $\mathbf{15}$    & $\mathbf{20}$     & $\mathbf{25}$     & $\mathbf{30}$     & $\mathbf{35}$     & $\mathbf{40}$     \\
\hline
$\mathbf{1^2}$      & $\!\!1.22$     & $1.28  $     &$ 1.31  $&$ 1.35  $&$ 1.39   $&$ 1.45   $&$ 1.50   $&$ 1.54   $&$ 1.58$   \\
$\mathbf{4^2}$      & $\!\!1.37$     & $1.74  $    &$ 2.22  $&$ 2.70  $&$ 3.19   $&$ 3.67   $&$ 4.06   $&$ 4.51   $&$ 4.97$   \\
$\mathbf{8^2}$      & $\!\!1.72$     & $3.43  $    &$ 5.18  $&$ 7.18  $&$ 8.91   $&$ 10.72  $&$ 12.54  $&$ 14.35  $&$ 16.96$  \\
$\mathbf{\!\!12^2}$ & $\!\!2.45$     & $6.59  $    &$\! 12.36 $&$ \!16.32 $&$ \!21.84  $&$ 26.34  $&$ 29.12  $&$ 34.12  $&$ 37.43$  \\
$\mathbf{\!\!16^2}$ & $\!\!3.34$     & $\!10.11 $  &$\! 18.15 $&$ \!23.62 $&$ \!33.68  $&$ 39.24  $&$ 49.46  $&$ 53.24  $&$ 68.60$  \\
$\mathbf{\!\!20^2}$ & $\!\!4.65$     & $\!13.25 $  &$\! 27.46 $&$ \!37.58 $&$ \!52.63  $&$ 63.24  $&$ 76.67  $&$ 83.14  $&$ 92.32$  \\
$\mathbf{\!\!24^2}$ & $\!\!5.92$     & $\!21.62 $  &$\! 39.65 $&$ \!55.28 $&$ \!75.68  $&$ 91.35  $&$ \!\!109.43 $&$ \!\!113.43 $&$ \!\!118.53$ \\
$\mathbf{\!\!28^2}$ & $\!\!7.46$     & $\!28.12 $  &$\! 52.38 $&$ \!83.03 $&$ \!\!102.71 $&$ \!\!134.62 $&$ \!\!150.87 $&$ \!\!189.41 $&$ \!\!201.99$ \\
$\mathbf{\!\!32^2}$ & $\!\!9.36$     & $\!35.89 $  &$\! 69.81 $&$ \!98.60 $&$ \!\!116.35 $&$ \!\!148.35 $&$ \!\!185.65 $&$ \!\!225.32 $&$ \!\!257.34$ \\
\hline
\hline
\!\!\!$p$\verb|\|$k$ & $\mathbf{1}$    & $\mathbf{5}$     & $\mathbf{10}$    & $\mathbf{15}$    & $\mathbf{20}$     & $\mathbf{25}$     & $\mathbf{30}$     & $\mathbf{35}$     & $\mathbf{40}$     \\
\hline
$\mathbf{1^2}$  & $\!\!0.02$  & $0.08$  & $0.18$  & $0.22$   & $0.47$   & $0.8$    & $1.08$   & $1.22$   & $2.33$   \\
$\mathbf{4^2}$  & $\!\!0.18$  & $0.61$  & $1.54$  & $2.75$   & $5.18$   & $7.82$   & $\!12.76$  & $\!18.12$  & $\!26.53$  \\
$\mathbf{8^2}$  & $\!\!0.58$  & $2.11$  & $5.32$  & $\!10.98$  & $\!18.95$  & $\!32.34$  & $\!49.56$  & $\!71.31$  & $\!97.61$  \\
$\mathbf{\!\!12^2}$ & $\!\!1.53$  & $7.18$  & $\!12.34$ & $\!22.98$  & $\!34.17$  & $\!50.37$  & $\!\!69.25$  & $\!95.34$  & $\!\!189.18$ \\
$\mathbf{\!\!16^2}$ & $\!\!2.94$  & $\!11.75$ & $\!20.48$ & $\!31.63$  & $\!48.23$  & $\!93.24$  & $\!\!123.85$ & $\!\!165.76$ & $\!\!323.71$ \\
$\mathbf{\!\!20^2}$ & $\!\!4.94$  & $\!20.13$ & $\!33.53$ & $\!64.55$  & $\!\!101.34$ & $\!\!143.71$ & $\!\!198.35$ & $\!\!256.37$ & $\!\!358.35$ \\
$\mathbf{\!\!24^2}$ & $\!\!8.31$  & $\!28.74$ & $\!46.08$ & $\!92.96$  & $\!\!144.79$ & $\!\!203.84$ & $\!\!283.63$ & $\!\!372.54$ & $\!\!517.37$ \\
$\mathbf{\!\!28^2}$ & $\!\!\!12.83$ & $\!34.97$ & $\!69.77$ & $\!\!125.74$ & $\!\!197.36$ & $\!\!279.56$ & $\!\!387.19$ & $\!\!503.87$ & $\!\!697.64$ \\
$\mathbf{\!\!32^2}$ & $\!\!\!20.23$ & $\!51.54$ & $\!98.14$ & $\!\!165.25$ & $\!\!257.23$ & $\!\!367.95$ & $\!\!506.45$ & $\!\!658.32$ & $\!\!921.61$ \\
\hline
\hline
\end{tabular}
\label{tab:no_pruning_module}
\end{table}

\subsection{File Size and Time Cost}
\label{sec:timecost}
\textbf{File size:} in MARVEL, the pruning mechanism (PM) is designed to reduce the file sizes of vectorized outputs. To evaluate the effectiveness of PM, we randomly collect 50 raster inputs and generate 100 vectored samples under different settings of $p$ and $k$. Next, we measure the performance of PM in two aspects, the reduced file sizes and the influences on visual similarity. As shown in Figure \ref{fig:filesize}, the PM effectively reduces the file sizes by an average of $50.53\%$ when ${k}=40$. Moreover, the observed results show that the reduced file sizes are proportional to $k$, since the produced repeated and erroneous strokes are increasing with $k$ and are pruned by PM.

In Figure \ref{fig:visualsimilarity}, we display the measured average visual similarities between raster input and vectorized outputs, when $p=32\times32$ and $k=40$. The scatter and box charts show that the pruning mechanism can preserve and even improve the visual similarity when reducing file sizes, since the mechanism removes the error stroke lines.
To summarize, the proposed PM significantly reduces the file sizes of vectorized results without compromising the visual similarity.

\textbf{Time cost:} we evaluate the time costs of our vectorization without PM (only DRL processing) and the time costs of the PM processing. Under different settings of  ${p}$ and ${k}$, the calculated average time costs (in seconds) are shown in Table \ref{tab:no_pruning_module}, which shows that our time costs are increasing with $p$ and $k$, and the maximum vectorization time is the acceptable few minutes. Empirically, under the condition of using PM, setting $\{p=16^2$, $k=20\}$ is enough to produce results with high accuracy (average 81.91 seconds), and setting $\{p=32^2$, $k=20\}$ can further vectorize mangas with extremely complex structures (average 153.97 seconds). The above time costs are computed in single-thread mode, and using the multithreading mode can further save the run time.

\textbf{Memory usage and rendering time:} we evaluate the memory usage and render time of our results on different devices. The average test results of 100 random samples are shown in Table \ref{tab:memoryusage}. Experimental results show that our results are within an acceptable range in terms of render time and memory usage in various computers and mobile devices, which can satisfy different usage scenarios.

\subsection{Evaluation of vectorizing images with different resolutions}
We conduct experiments to evaluate our performance in visual effects, file sizes, and accuracies, when vectorizing mangas with different resolutions. 

\begin{figure*}[t]
\centering
\includegraphics[width=7.2 in]{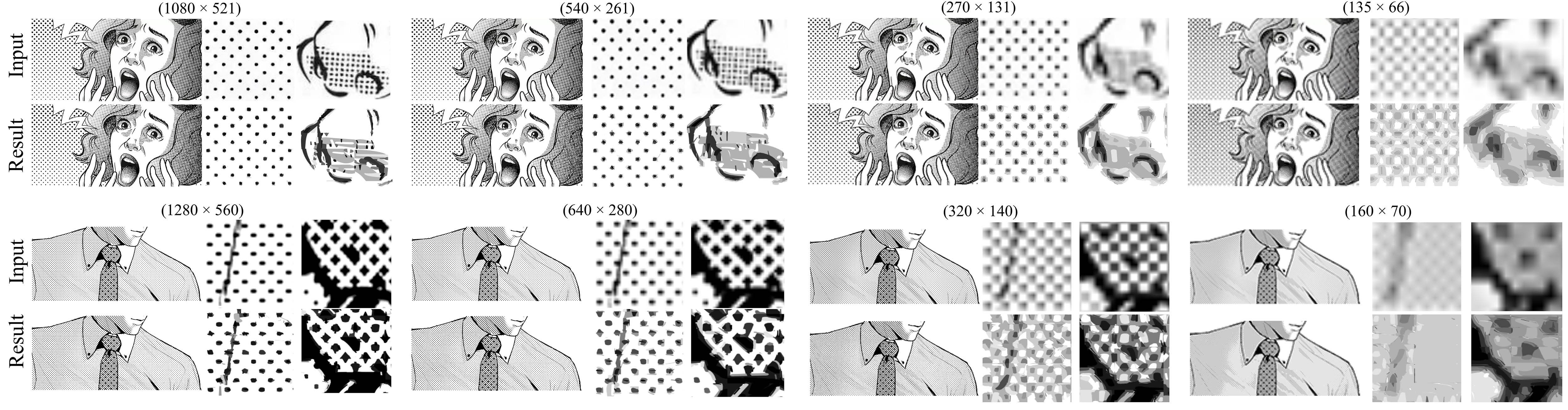}
\caption{Vectorized  screentones of input images with different resolutions.}
\label{fig:resolution_visual}
\end{figure*}

\begin{figure}[t]
\centering
\includegraphics[width=3.5 in]{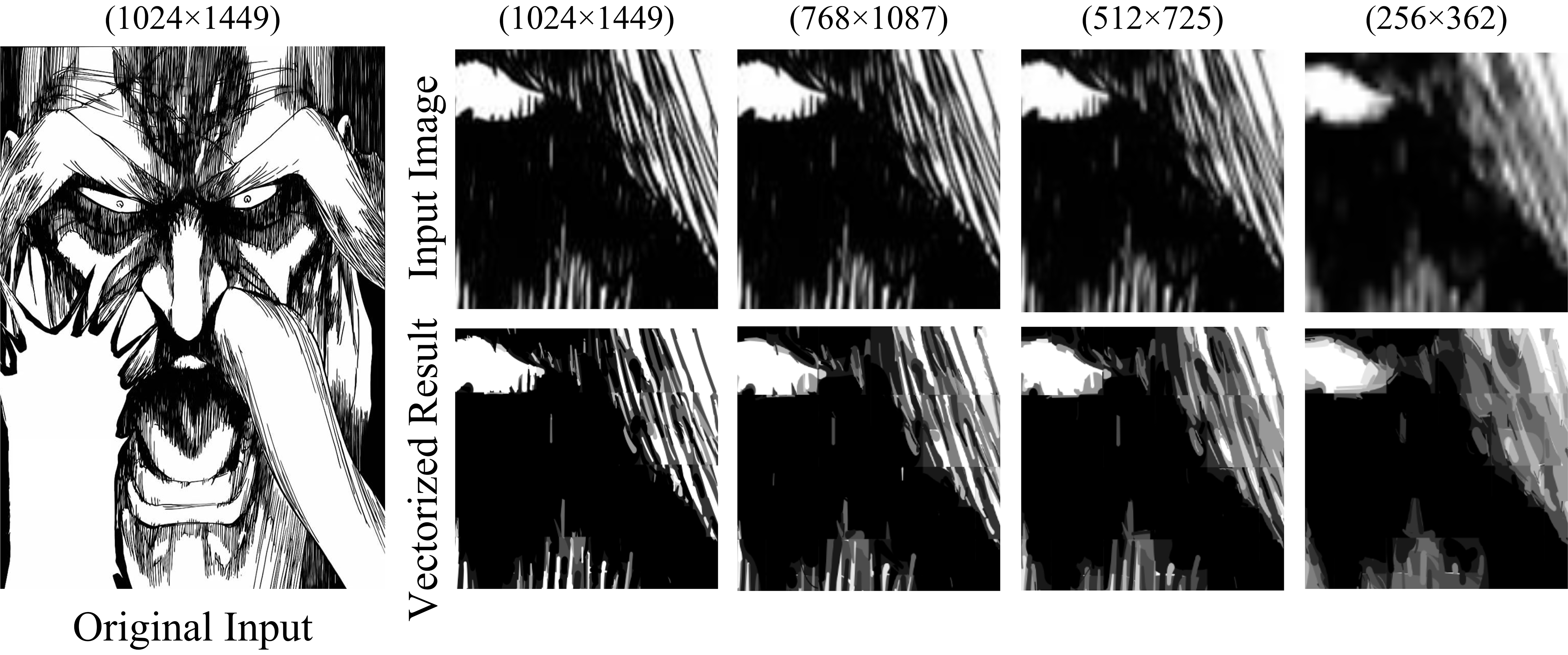}
\caption{Enlarged view of vectorized results of input images with different resolutions. The objective measurement results are summarized in Table \ref{tab:accuracy_diff_resolution}.}
\label{fig:enlarge_resolution}
\end{figure}

\textbf{Visual perception and screentone:} we evaluate our vectorized results of input mangas with different resolutions. As shown in Figure \ref{fig:resolution_visual} and Figure \ref{fig:enlarge_resolution}, for a target manga, we first resize it to different resolutions by CSI (cubic spline interpolation \cite{mckinley1998cubic}) , and then generate the corresponding vectorized results (p=$16^2$, k=$20$). We can observe that our method produces blurry  manga screentones in low image resolution, since our goal is to make results similar to the input mangas, and it is a limitation of our method that cannot reconstruct manga screentones. Therefore, for reconstructing screentones, MMV \cite{TVCG2016manga} performs better than our method.

\begin{figure}[t]
\centering
\includegraphics[width=3.2 in]{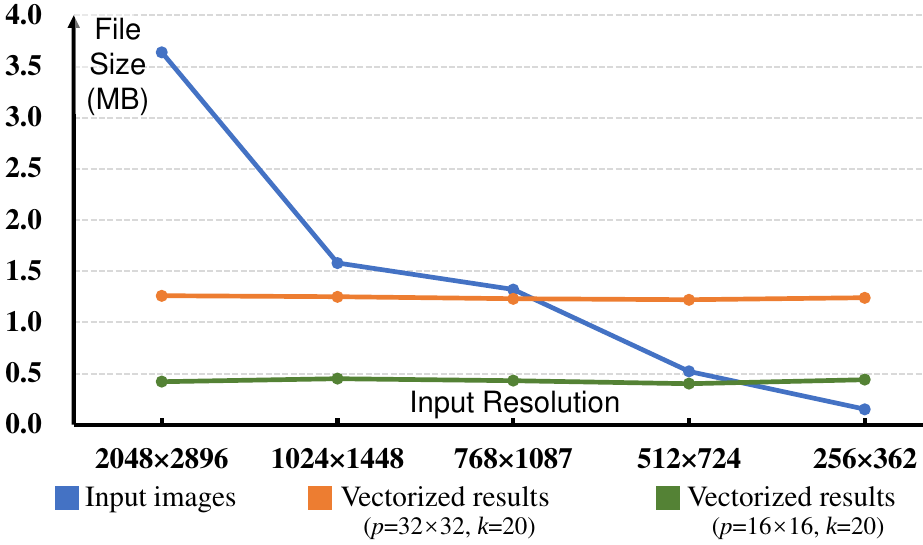}
\caption{Evaluation of reduced file sizes between input images and our vectorized results, under different input resolutions. Combined with Figure \ref{fig:filesize}, the input image resolutions do not influence our vectorized file sizes which are mainly influenced by the numbers of generated strokes $k$ and patches $p$. For input images with low resolution (e.g., 256$\times$362), we can further set smaller $p$ and $k$ to reduce vectorized file sizes.}
\label{fig:filesize_resolutioin}
\end{figure}

\textbf{File size:}  Figure \ref{fig:filesize_resolutioin} shows the average file sizes between input images and our vectorized results, under different input resolutions. Each shown file size is the average result computed by 50 random cases, and each image is in PNG format. Combined with Figure \ref{fig:filesize}, we obverse that the resolutions of input images do not influence the sizes of our vectorized results which are mainly influenced by the numbers of generated strokes $k$ and patches $p$. 
Our method can save file sizes when setting $p=32^2$ and $k=20$ for high accuracy (saved about $21.4\%$ file size when input resolutions are $1024$$\times$$1448$), and $p=16^2$ and $k=20$ for common accuracy (saved about $59.3\%$ file sizes when input resolutions are $768\times1087$). Moreover, for input images with low resolution (e.g., $256$$\times$$362$), we can further set smaller $p$ and $k$ to reduce vectorized file sizes.

\textbf{Vectorization accuracy:} first, we randomly select 50 A4 pages of high-resolution input mangas (normalized to $1024$$\times$$1448$) . Let $M_0$ indicate the original high-resolution input manga, we downsample $M_0$ to $M_1$ ($768$$\times$$1087$), $M_2$ ($512$$\times$$724$), and $M_3$ ($256$$\times$$362$) by CSI \cite{mckinley1998cubic}, respectively. Then, we vectorize $\{M_0$, $M_1$, $M_2$, $M_3\}$ to $\{V_0$, $V_1$, $V_2$, $V_3\}$ (p=$32^2$, k=$20$), and conduct two evaluations $\mathrm{D_{img}}$ and $\mathrm{D_{vec}}$. In $\mathrm{D_{img}}$ (or $\mathrm{D_{vec}}$), we measure differences between $M_0$ and $M_\delta$ (or differences between $M_0$ and $\mathcal{R}(V_\delta)$), where $\delta$$\in$$\{0,1,2,3\}$, and the function $\mathcal{R}(V_\delta)$ is utilized to rasterize $\mathcal{V}_\delta$ to images by CairoSVG \cite{cairosvg}. 

The average results are summarized in Table \ref{tab:accuracy_diff_resolution}, and some magnification examples are shown in Figure \ref{fig:resolution_visual} and Figure \ref{fig:enlarge_resolution}. We observe that our vectorization accuracies between $\mathcal{R}(V_\delta)$ and $M_0$ are directly proportional to the similarities between $M_\delta$ and $M_0$. Since our method aims to preserve high accuracy between inputs and outputs, thus our method cannot reconstruct high-resolution images by vectorizing the downsampled low-resolution images.

\begin{table}[t]
\centering
\caption{Average vectorization accuracy at different resolutions.}
\scriptsize
\begin{tabular}{|m{0.5cm}<{\centering}m{0.8cm}<{\centering}|c|c|c|c|}
\hline
\multicolumn{2}{|m{1.5cm}<{\centering}|}{$\!\!\!\!\!\!$\textbf{Index}$\setminus$\textbf{Resolution}}                   & $1024$$\times$$1448$ & $768$$\times$$1087$ & $512$$\times$$724$ & $256$$\times$$362$ \\ \hline
\multicolumn{1}{|m{0.5cm}<{\centering}|}{\multirow{2}{*}{$\mathrm{D_{img}}$}} & {MSE}  & 0.0000         & 0.0242        & 0.0929       & 0.1410       \\ \cline{2-6} 
\multicolumn{1}{|m{0.5cm}<{\centering}|}{}                           & {SSIM} & 1.0000         & 0.9807        & 0.9170       & 0.6897       \\ \hline
\multicolumn{1}{|m{0.5cm}<{\centering}|}{\multirow{2}{*}{$\mathrm{D_{vec}}$}} & {MSE}  & 0.0813        & 0.0826         & 0.1232     & 0.1804      \\ \cline{2-6} 
\multicolumn{1}{|m{0.5cm}<{\centering}|}{}                           & {SSIM} & 0.9252        & 0.8842       & 0.7693      & 0.5978      \\ \hline
\end{tabular}
\label{tab:accuracy_diff_resolution}
\end{table}

\renewcommand\arraystretch{1.2}
\begin{table}[t]
\centering
\caption{Memory usage and rendering time of our vectorized results under different situations.}
\scriptsize
\begin{tabular}{m{0.8cm}<{\centering}m{0.8cm}<{\centering}|m{0.8cm}<{\centering}|m{1cm}<{\centering}|m{0.8cm}<{\centering}|m{0.8cm}<{\centering}|m{0.7cm}<{\centering}}
\hline
\hline
\multicolumn{2}{c|}{\textbf{\makecell[c]{Device}}$\!\!$}                                                                & $\!\!$\textbf{\makecell[c]{$\!\!\!$Operating\\$\!\!\!$System}}              & \textbf{PDF Reader}    & \textbf{File Size}                                                       & \textbf{Memory Usage} & \multicolumn{1}{m{0.6cm}<{\centering}}{\textbf{\makecell[c]{$\!\!\!$Render\\$\!\!\!$Time}}} \\
\hline
\hline
\multicolumn{1}{m{0.8cm}<{\centering}|}{\multirow{6}{*}{\vspace{-0.5cm}$\!\!$\makecell[c]{Mobile\\Phone}}} & \multirow{3}{*}{\vspace{-0.5cm}$\!\!$\makecell[c]{Iphone 12}}    & \multirow{3}{*}{\vspace{-0.5cm}$\!\!$IOS}          & Chrome        & \multirow{12}{*}{\vspace{-1.1cm}$\!\!\!$\makecell[c]{648.38KB \\ ($p=16^2$,\\ $k=40$) }} & -            & 0.52s                               \\ \cline{4-4} \cline{6-7}
\multicolumn{1}{c|}{}                              &                                       &                               & Adobe Acrobat &                                                                                                       & -            & 0.44s                               \\ \cline{4-4} \cline{6-7}
\multicolumn{1}{c|}{}                              &                                       &                               & Foxit Reader  &                                                                                                       & -            & 0.63s                               \\ \cline{2-4} \cline{6-7}
\multicolumn{1}{l|}{}                              & \multirow{3}{*}{\vspace{-0.5cm}$\!\!$\makecell[c]{HUAWEI\\Mate 40}}       & \multirow{3}{*}{\vspace{-0.5cm}$\!\!$\makecell[c]{EMUI/\\Android}} & Chrome        &                                                                                                       & -            & 0.48s                               \\ \cline{4-4} \cline{6-7}
\multicolumn{1}{l|}{}                              &                                       &                               & Adobe Acrobat &                                                                                                       & -            & 0.65s                               \\ \cline{4-4} \cline{6-7}
\multicolumn{1}{l|}{}                              &                                       &                               & Foxit Reader  &                                                                                                       & -            & 0.55s                               \\ \cline{1-4} \cline{6-7}
\multicolumn{1}{m{0.8cm}<{\centering}|}{\multirow{6}{*}{\vspace{-0.5cm}$\!\!$\makecell[c]{Computer}}}     & \multirow{3}{*}{\vspace{-0.5cm}$\!\!$\makecell[c]{Dell\\Alienware\\m17R5}} & \multirow{3}{*}{\vspace{-0.5cm}$\!\!$Windows}      & Chrome        &                                                                                                       & 11.3MB       & 0.45s                               \\ \cline{4-4} \cline{6-7}
\multicolumn{1}{l|}{}                              &                                       &                               & Adobe Acrobat &                                                                                                       & 28.4MB       & 0.65s                               \\ \cline{4-4} \cline{6-7}
\multicolumn{1}{l|}{}                              &                                       &                               & Foxit Reader  &                                                                                                       & 21.3MB       & 0.58s                               \\ \cline{2-4} \cline{6-7}
\multicolumn{1}{l|}{}                              & \multirow{3}{*}{\vspace{-0.5cm}$\!\!$Mac book}             & \multirow{3}{*}{\vspace{-0.5cm}$\!\!$IOS}          & Chrome        &                                                                                                       & 10.9MB       & 0.42s                               \\ \cline{4-4} \cline{6-7}
\multicolumn{1}{l|}{}                              &                                       &                               & Adobe Acrobat &                                                                                                       & 27.9MB       & 0.53s                               \\ \cline{4-4} \cline{6-7}
\multicolumn{1}{l|}{}                              &                                       &                               & Foxit Reader  &                                                                                                       & 11.2MB       & 0.56s                               \\ \hline
\end{tabular}
\label{tab:memoryusage}
\end{table}

%

\begin{figure}[t]
\centering
\vspace{0.5cm}
\includegraphics[width=3.5 in]{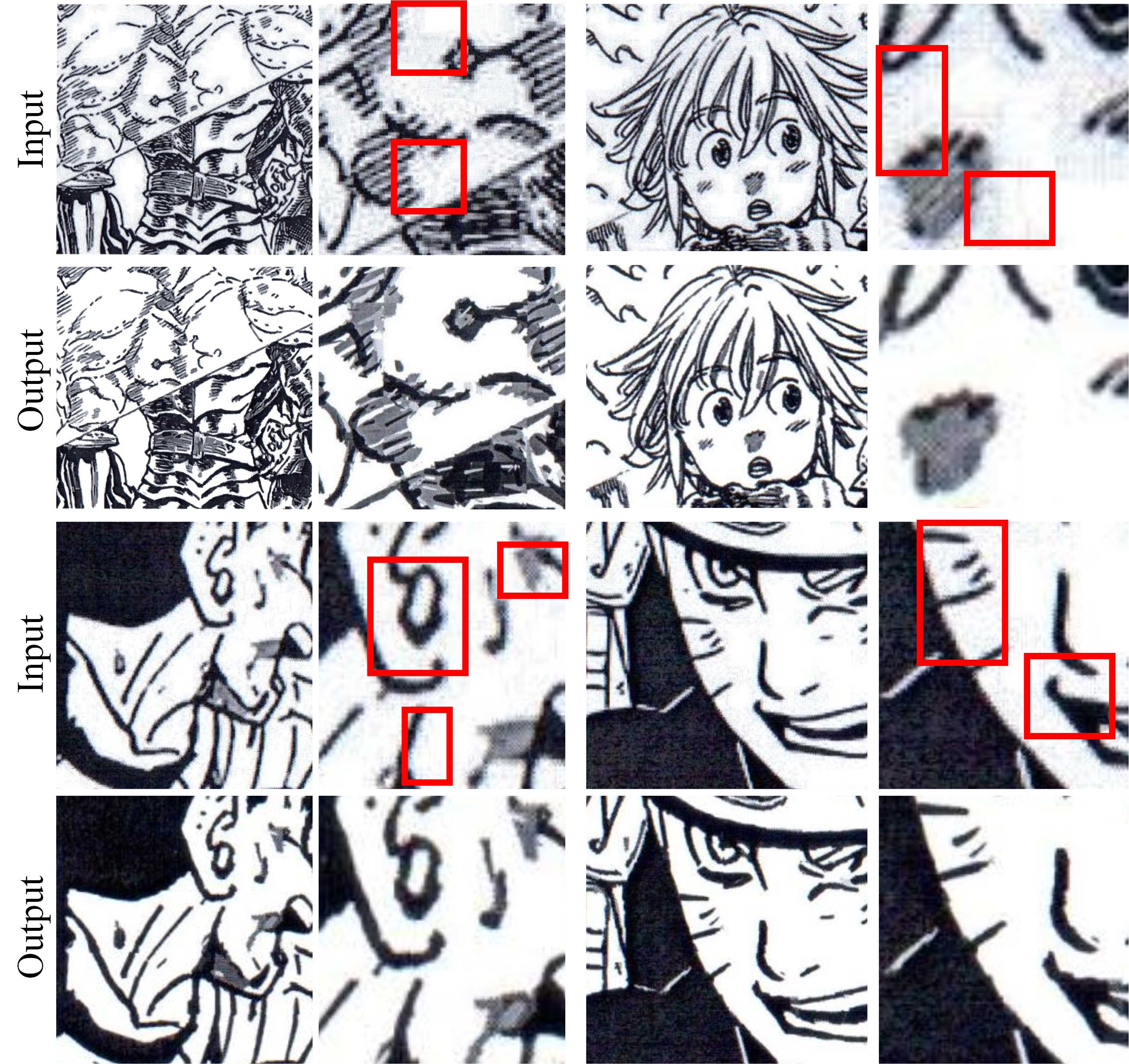}
\caption{Scan results of print versions in the enlarged view. By contrast, our vectorized outputs have cleaner colors and lines than raster inputs (red boxes), since indistinct and irregular pixels will be filtered when our model predicts stroke lines.}
\label{fig:print}
\end{figure}

\subsection{Evaluation of Print version}
To evaluate our vectorized results' quality in the print version, we select 20 paired samples (raster inputs and vectorized outputs) with extremely complex textures. Next, each sample is printed on A4 paper (210  mm$\times$297 mm)  and then scanned. The printer and scanner used in experiments is HP LaserJet M1535dnf MFP.

Figure \ref{fig:printsample} and Figure \ref{fig:print} show the scan results of print versions. By contrast, our vectorized results significantly preserve the visual details of raster inputs, even in the enlarged view.
In addition, the colors and lines of our vectorized outputs are even cleaner than raster inputs, since some indistinct and irregular pixels (e.g., artifacts in red boxes of Figure \ref{fig:print}) will be filtered when our model predicts stroke lines.

%

\section{Discussion and Limitations}

\begin{figure}[t]
\vspace{0.5cm}
\centering
\includegraphics[width=3.2 in]{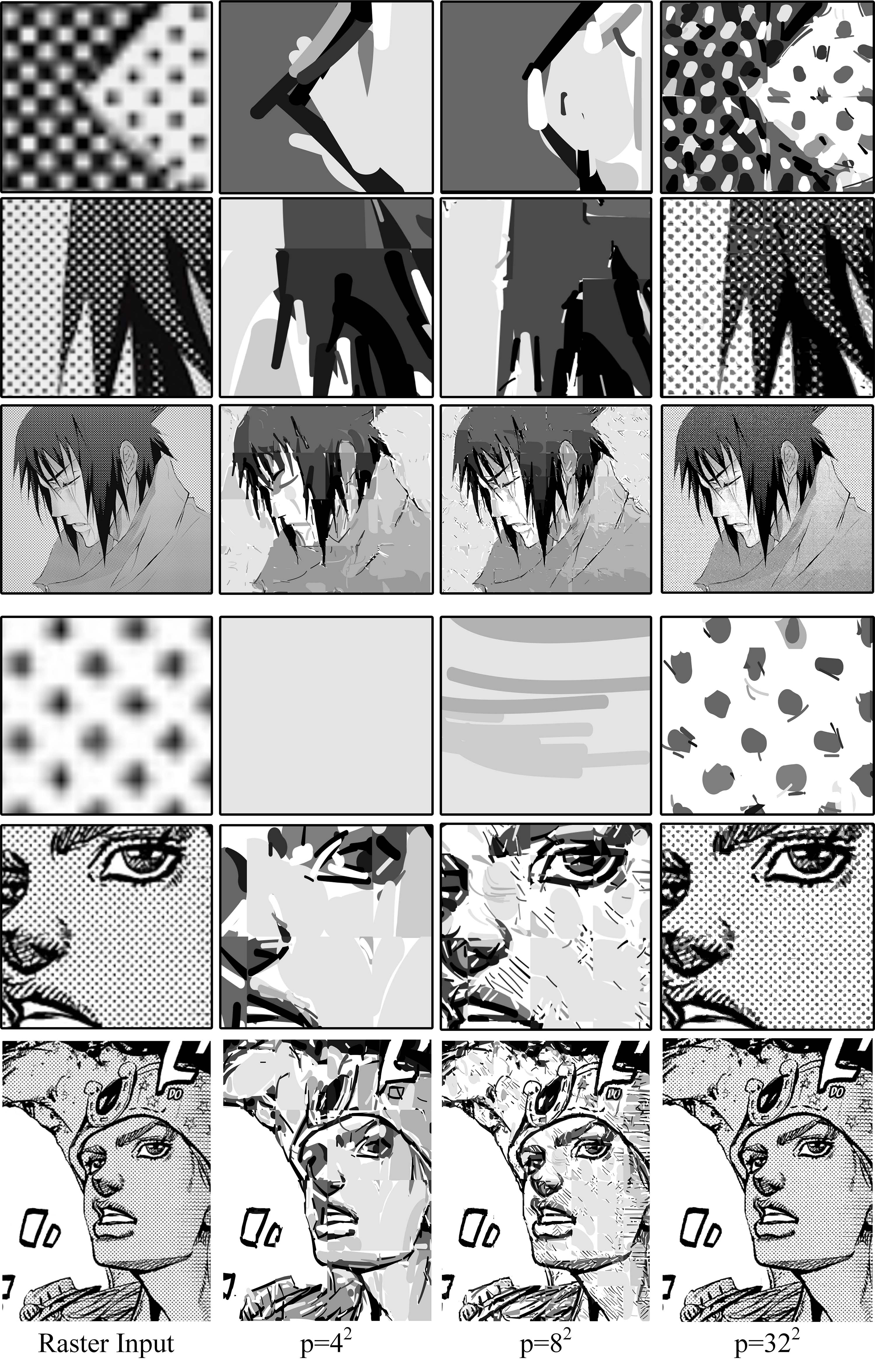}
\caption{Vectorized results and enlarged views of manga screentones under different settings of $p$, when ${k}$$=$$20$. }
\label{fig:screentones}
\end{figure}

\begin{figure}[t]
\centering
\includegraphics[width=3.5 in]{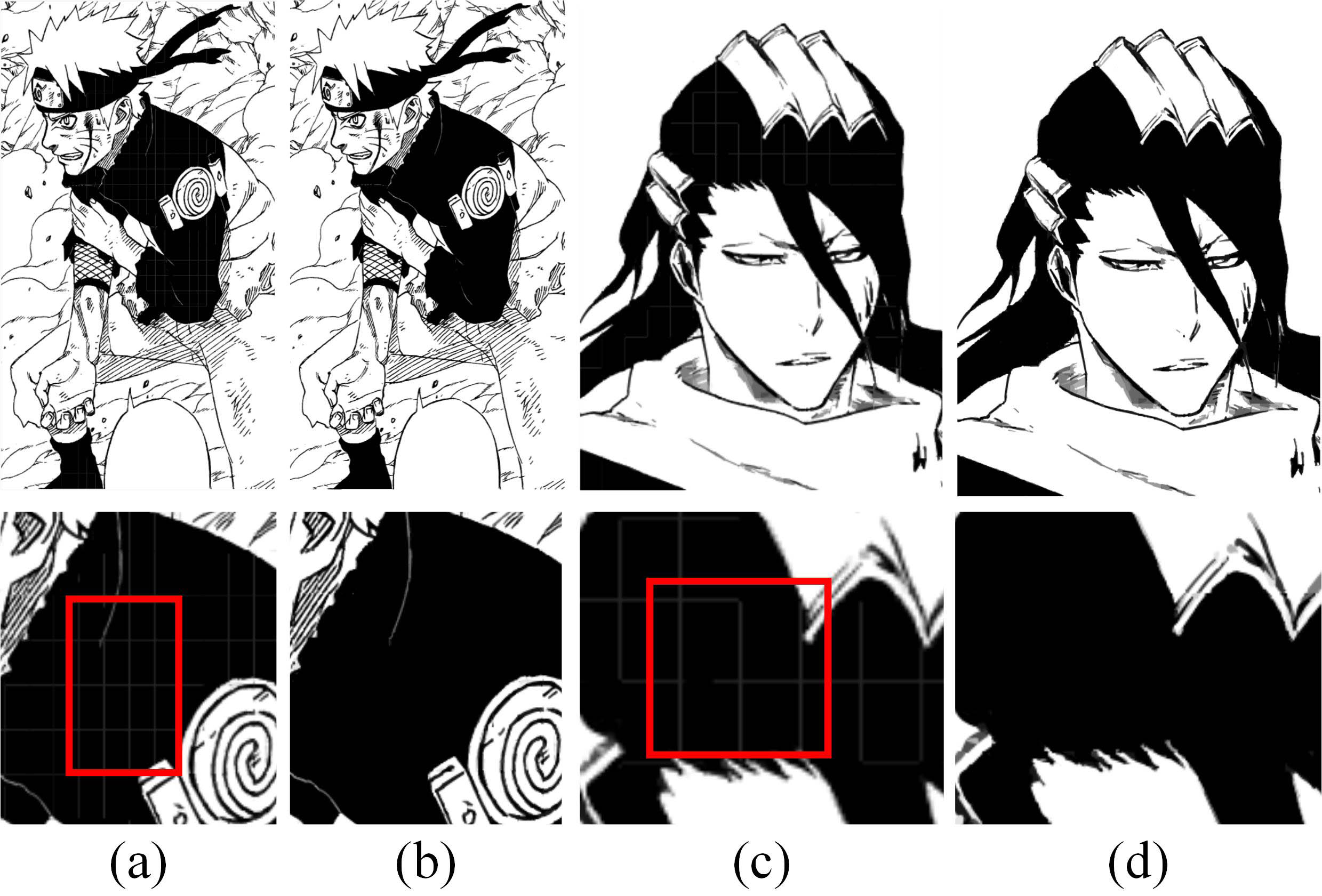}
\caption{Screenshots of our results in different PDF readers. (a)(c) Screenshots in SumatraPDF. (b)(d) Screenshots in Chrome. Our results may be rendered visually unpleasant by some PDF readers (e.g., SumatraPDF) which incurs white gaps among square patches (red boxes). This issue can be solved by zooming in or utilizing other PDF readers (e.g., Chrome, Foxit Reader), and these different rendering effects will not affect the visual quality of the printed version.}  
\label{fig:limitation}
\end{figure}

\subsection{Discussion}
\textbf{Machine learning of manga strokes:} the previous literature MMV \cite{TVCG2016manga} proposes a stroke-based concept for future work that develops a machine-learning mechanism to automatically add extra manga strokes after deformation. Our work is similar to the stroke-based concept proposed by MMV \cite{TVCG2016manga}. By comparison, our method is implemented by a tailored deep reinforcement learning framework, and further proposes an accuracy-oriented reward function and action model of construct stroke lines. In addition, we decompose an entire image into strokes, rather than extracting extra strokes

\textbf{Screentones:} screentone \cite{TOG2021r3} is a technique for applying textures and shades to manga, used as an alternative to hatching. In our method, we vectorize a screentone region like vectorizing a general pixel region. Specifically, as shown in Figure \ref{fig:screentones}, the parameter $p$ controls the detail level of our vectorized outputs. When $p$ is large (e.g., $p=32^2$), the textures and dots in screentones are obvious and clean. When p is small (e.g., $p$$\leqslant$$8$), screentones will be vectorized to a region with an average color.

 \textbf{Usage scenario:} our vectorization method can translate raster mangas to vector graphics with preserving high similarities in both electric and print versions. Moreover, for showing high-resolution contents, the vectorized mangas have smaller file sizes than raster images. Hence, our method can be applied to store raster mangas in a resolution-independent manner, which is readily displayed on digital devices with different resolutions.

In addition, besides our current usage scenarios, our work presents an enlightening contribution to the application and research of learning-based vectorization. Specifically, in our current method, our vectorized results are composed of many vector primitives based on B\'{e}zier curves (as shown in Figure \ref{fig:fiting-module}), which limits our performance in shape retrieval and content editing. In our future work, we will explore an algorithm that merges two B\'{e}zier primitives into one primitive in each step, to extract total vector paths. Then, our method can be applied to the same scenarios (e.g., shape retrieval, content editing) as other vectorization methods (algorithm-based or learning-based).

 \textbf{Processable vector parameter:} for the learning-based methods we have compared, Im2Vec \cite{2021im2vec}, SVG-VAE \cite{svgvae}, and DeepSVG \cite{deepsvg} vectorize an entire image in a one-step manner that limits the number of predictable vector parameters (i.e., several B\'{e}zier curves). In addition, LIVE \cite{ma2022towards} is a special learning-based method that fits curves in a stepwise  manner using a neural optimizer without training any model, but the method requires hundreds of iterations to fit each vector curve. By contrast, our method combines the merits of learning-based and stepwise methods, which not only adopts the trained model to fast predict strokes, but also constructs a DRL framework that can infinitely add stroke lines to correct inaccurate areas in a stepwise manner.

 \textbf{Patch Segmentation:} MARVEL first segments a raster image into small patches and then vectorizes each patch, and this manner has two advantages. First, it allows and benefits users to control the levels of vectorization detail. Second, the hundreds of patches can be considered as one input batch of the trained model. Then, the model can vectorize all images in the input batch simultaneously, which significantly increases the vectorization speed. There may be introduced some boundary artifacts, but these artifacts are almost ignorable when setting a large $p$.

\subsection{Limitation}
\textbf{Visual effect in different readers:} as shown in Figure \ref{fig:limitation}, our results may be rendered visually unpleasant by some PDF readers (e.g., SumatraPDF), which causes white gaps among square patches (marked in red boxes). This issue can be solved by zooming in or utilizing other PDF readers (e.g., Chrome, Foxit Reader), and these different rendering effects will not affect the visual quality of the printed version as shown in Figure \ref{fig:printsample}.

 \textbf{Shape editing:} our MARVEL is a primitive-based method that represents the global vector graphic as a collection of vector primitives. This is much different from other methods that mainly predict some key vector paths (e.g., \cite{svgvae,deepsvg,2021im2vec, tcsvtliu2015making, tcsvtyao2016resolution}), and thus our method is not convenient in shape editing.

\section{Conclusion}

In this paper, we propose MARVEL, a primitive-wise approach for vectorizing raster mangas by DRL. MARVEL introduces a novel perspective that regards an entire manga as a collection of primitives\textemdash stroke lines that can be decomposed from the target image for achieving accurate vectorization.
Extensive subjective and objective experiments have proved the effectiveness of our improvements, and have shown that compared with both algorithm-based and learning-based methods, our MARVEL can produce impressive results and has reached the state-of-the-art level.

In the future, we aim to explore a sliding window mechanism to avoid the isolation of patches, and explore a stroke fusion mechanism to obtain key paths for achieving convenient shape editing.

\bibliographystyle{IEEEtran}
\bibliography{ref}

\begin{thebibliography}{10}
\providecommand{\url}[1]{#1}
\csname url@samestyle\endcsname
\providecommand{\newblock}{\relax}
\providecommand{\bibinfo}[2]{#2}
\providecommand{\BIBentrySTDinterwordspacing}{\spaceskip=0pt\relax}
\providecommand{\BIBentryALTinterwordstretchfactor}{4}
\providecommand{\BIBentryALTinterwordspacing}{\spaceskip=\fontdimen2\font plus
\BIBentryALTinterwordstretchfactor\fontdimen3\font minus
  \fontdimen4\font\relax}
\providecommand{\BIBforeignlanguage}[2]{{%
\expandafter\ifx\csname l@#1\endcsname\relax
\typeout{** WARNING: IEEEtran.bst: No hyphenation pattern has been}%
\typeout{** loaded for the language `#1'. Using the pattern for}%
\typeout{** the default language instead.}%
\else
\language=\csname l@#1\endcsname
\fi
#2}}
\providecommand{\BIBdecl}{\relax}
\BIBdecl

\bibitem{tcsvtzhu2012object}
Z.-Y. Zhu, S.~Zhang, S.-C. Chan, and H.-Y. Shum, ``Object-based rendering and
  3-d reconstruction using a moveable image-based system,'' \emph{IEEE
  Transactions on Circuits and Systems for Video Technology (TCSVT)}, vol.~22,
  no.~10, pp. 1405--1419, 2012.

\bibitem{tog2008}
A.~Orzan, A.~Bousseau, H.~Winnem{\"o}ller, P.~Barla, J.~Thollot, and
  D.~Salesin, ``Diffusion curves: a vector representation for smooth-shaded
  images,'' \emph{ACM Transactions on Graphics (TOG)}, vol.~27, no.~3, pp.
  1--8, 2008.

\bibitem{tvcg2009}
S.-H. Zhang, T.~Chen, Y.-F. Zhang, S.-M. Hu, and R.~R. Martin, ``Vectorizing
  cartoon animations,'' \emph{IEEE Transactions on Visualization and Computer
  Graphics (TVCG)}, vol.~15, no.~4, pp. 618--629, 2009.

\bibitem{depixelizing}
J.~Kopf and D.~Lischinski, ``Depixelizing pixel art,'' in \emph{ACM SIGGRAPH},
  2011, pp. 1--8.

\bibitem{line2019}
M.~Bessmeltsev and J.~Solomon, ``Vectorization of line drawings via polyvector
  fields,'' \emph{ACM Transactions on Graphics (TOG)}, vol.~38, no.~1, pp.
  1--12, 2019.

\bibitem{tog2013cartoon}
G.~Noris, A.~Hornung, R.~W. Sumner, M.~Simmons, and M.~Gross, ``Topology-driven
  vectorization of clean line drawings,'' \emph{ACM Transactions on Graphics
  (TOG)}, vol.~32, no.~1, pp. 1--11, 2013.

\bibitem{ardeco2006}
G.~Lecot and B.~Levy, ``Ardeco: Automatic region detection and conversion,'' in
  \emph{Eurographics Symposium on Rendering-EGSR}, 2006, pp. 349--360.

\bibitem{Donati_2017_ICCV}
L.~Donati, S.~Cesano, and A.~Prati, ``An accurate system for fashion hand-drawn
  sketches vectorization,'' in \emph{Proceedings of the IEEE International
  Conference on Computer Vision (ICCV) Workshops}, Oct 2017.

\bibitem{dominici2020polyfit}
E.~A. Dominici, N.~Schertler, J.~Griffin, S.~Hoshyari, L.~Sigal, and
  A.~Sheffer, ``Polyfit: perception-aligned vectorization of raster clip-art
  via intermediate polygonal fitting,'' \emph{ACM Transactions on Graphics
  (TOG)}, vol.~39, no.~4, pp. 77--1, 2020.

\bibitem{tcsvtding2019bilinear}
Y.~Ding, W.~K. Wong, Z.~Lai, and Z.~Zhang, ``Bilinear supervised hashing based
  on 2d image features,'' \emph{IEEE Transactions on Circuits and Systems for
  Video Technology (TCSVT)}, vol.~30, no.~2, pp. 590--602, 2019.

\bibitem{tcsvtyu2019multimodal}
J.~Yu, J.~Li, Z.~Yu, and Q.~Huang, ``Multimodal transformer with multi-view
  visual representation for image captioning,'' \emph{IEEE Transactions on
  Circuits and Systems for Video Technology (TCSVT)}, vol.~30, no.~12, pp.
  4467--4480, 2019.

\bibitem{egiazarian2020deep}
V.~Egiazarian, O.~Voynov, A.~Artemov, D.~Volkhonskiy, A.~Safin, M.~Taktasheva,
  D.~Zorin, and E.~Burnaev, ``Deep vectorization of technical drawings,'' in
  \emph{European Conference on Computer Vision (ECCV)}.\hskip 1em plus 0.5em
  minus 0.4em\relax Springer, 2020, pp. 582--598.

\bibitem{2021im2vec}
P.~Reddy, M.~Gharbi, M.~Lukac, and N.~J. Mitra, ``Im2vec: Synthesizing vector
  graphics without vector supervision,'' in \emph{Proceedings of the IEEE/CVF
  International Conference on Computer Vision (CVPR)}, 2021, pp. 7342--7351.

\bibitem{gao2019deepspline}
J.~Gao, C.~Tang, V.~Ganapathi-Subramanian, J.~Huang, H.~Su, and L.~J. Guibas,
  ``Deepspline: Data-driven reconstruction of parametric curves and surfaces,''
  \emph{arXiv:1901.03781}, 2019.

\bibitem{sinha2017surfnet}
A.~Sinha, A.~Unmesh, Q.~Huang, and K.~Ramani, ``Surfnet: Generating 3d shape
  surfaces using deep residual networks,'' in \emph{Proceedings of the IEEE/CVF
  International Conference on Computer Vision (CVPR)}, 2017, pp. 6040--6049.

\bibitem{Liu_2017_ICCV}
C.~Liu, J.~Wu, P.~Kohli, and Y.~Furukawa, ``Raster-to-vector: Revisiting
  floorplan transformation,'' in \emph{Proceedings of the IEEE International
  Conference on Computer Vision (ICCV)}, Oct 2017.

\bibitem{mangagan}
H.~Su, J.~Niu, X.~Liu, Q.~Li, J.~Cui, and J.~Wan, ``Mangagan: Unpaired
  photo-to-manga translation based on the methodology of manga drawing,'' in
  \emph{Proceedings of the AAAI Conference on Artificial Intelligence (AAAI)},
  vol.~35, no.~3, 2021, pp. 2611--2619.

\bibitem{artcoder}
H.~Su, J.~Niu, X.~Liu, Q.~Li, J.~Wan, M.~Xu, and T.~Ren, ``Artcoder: An
  end-to-end method for generating scanning-robust stylized qr codes,'' in
  \emph{Proceedings of the IEEE/CVF Conference on Computer Vision and Pattern
  Recognition (CVPR)}, 2021, pp. 2277--2286.

\bibitem{su2021q}
H.~Su, J.~Niu, X.~Liu, Q.~Li, J.~Wan, and M.~Xu, ``Q-art code: Generating
  scanning-robust art-style qr codes by deformable convolution,'' in
  \emph{Proceedings of the 29th ACM International Conference on Multimedia (ACM
  MM)}, 2021, pp. 722--730.

\bibitem{tcsvtsong2018deeply}
X.~Song, Y.~Dai, and X.~Qin, ``Deeply supervised depth map super-resolution as
  novel view synthesis,'' \emph{IEEE Transactions on circuits and systems for
  video technology (TCSVT)}, vol.~29, no.~8, pp. 2323--2336, 2018.

\bibitem{tcsvtzhang2014unified}
J.~Zhang, Y.~Cao, Z.-J. Zha, Z.~Zheng, C.~W. Chen, and Z.~Wang, ``A unified
  scheme for super-resolution and depth estimation from asymmetric stereoscopic
  video,'' \emph{IEEE Transactions on Circuits and Systems for Video Technology
  (TCSVT)}, vol.~26, no.~3, pp. 479--493, 2014.

\bibitem{tcsvtbo2022all}
Q.~Bo, W.~Ma, Y.-K. Lai, and H.~Zha, ``All-higher-stages-in adaptive context
  aggregation for semantic edge detection,'' \emph{IEEE Transactions on
  Circuits and Systems for Video Technology (TCSVT)}, 2022.

\bibitem{TCSVTliu2021cross}
Y.~Liu, Q.~Jia, X.~Fan, S.~Wang, S.~Ma, and W.~Gao, ``Cross-srn:
  Structure-preserving super-resolution network with cross convolution,''
  \emph{IEEE Transactions on Circuits and Systems for Video Technology
  (TCSVT)}, 2021.

\bibitem{guo2019deep}
Y.~Guo, Z.~Zhang, C.~Han, W.~Hu, C.~Li, and T.-T. Wong, ``Deep line drawing
  vectorization via line subdivision and topology reconstruction,'' in
  \emph{Computer Graphics Forum (CGF)}, vol.~38, no.~7.\hskip 1em plus 0.5em
  minus 0.4em\relax Wiley Online Library, 2019, pp. 81--90.

\bibitem{mo2021general}
H.~Mo, E.~Simo-Serra, C.~Gao, C.~Zou, and R.~Wang, ``General virtual sketching
  framework for vector line art,'' \emph{ACM Transactions on Graphics (TOG)},
  vol.~40, no.~4, pp. 1--14, 2021.

\bibitem{ma2022towards}
X.~Ma, Y.~Zhou, X.~Xu, B.~Sun, V.~Filev, N.~Orlov, Y.~Fu, and H.~Shi, ``Towards
  layer-wise image vectorization,'' in \emph{Proceedings of the IEEE/CVF
  Conference on Computer Vision and Pattern Recognition (CVPR)}, 2022, pp.
  16\,314--16\,323.

\bibitem{ha2017neural}
D.~Ha and D.~Eck, ``A neural representation of sketch drawings,''
  \emph{arXiv:1704.03477}, 2017.

\bibitem{ribeiro2020sketchformer}
L.~S.~F. Ribeiro, T.~Bui, J.~Collomosse, and M.~Ponti, ``Sketchformer:
  Transformer-based representation for sketched structure,'' in
  \emph{Proceedings of the IEEE/CVF Conference on Computer Vision and Pattern
  Recognition (CVPR)}, 2020, pp. 14\,153--14\,162.

\bibitem{svgvae}
R.~G. Lopes, D.~Ha, D.~Eck, and J.~Shlens, ``A learned representation for
  scalable vector graphics,'' in \emph{Proceedings of the IEEE/CVF
  International Conference on Computer Vision (CVPR)}, 2019, pp. 7930--7939.

\bibitem{deepsvg}
A.~Carlier, M.~Danelljan, A.~Alahi, and R.~Timofte, ``Deepsvg: A hierarchical
  generative network for vector graphics animation,'' \emph{Advances in Neural
  Information Processing Systems}, vol.~33, pp. 16\,351--16\,361, 2020.

\bibitem{azadi2018multi}
S.~Azadi, M.~Fisher, V.~G. Kim, Z.~Wang, E.~Shechtman, and T.~Darrell,
  ``Multi-content gan for few-shot font style transfer,'' in \emph{Proceedings
  of the IEEE conference on computer vision and pattern recognition (CVPR)},
  2018, pp. 7564--7573.

\bibitem{gao2019artistic}
Y.~Gao, Y.~Guo, Z.~Lian, Y.~Tang, and J.~Xiao, ``Artistic glyph image synthesis
  via one-stage few-shot learning,'' \emph{ACM Transactions on Graphics (TOG)},
  vol.~38, no.~6, pp. 1--12, 2019.

\bibitem{differentiable}
T.-M. Li, M.~Luk{\'a}{\v{c}}, M.~Gharbi, and J.~Ragan-Kelley, ``Differentiable
  vector graphics rasterization for editing and learning,'' \emph{ACM
  Transactions on Graphics (TOG)}, vol.~39, no.~6, pp. 1--15, 2020.

\bibitem{tog2017r1}
C.~Li, X.~Liu, and T.-T. Wong, ``Deep extraction of manga structural lines,''
  \emph{ACM Transactions on Graphics (TOG)}, vol.~36, no.~4, pp. 1--12, 2017.

\bibitem{ICCV2021r2}
M.~Xia, W.~Hu, X.~Liu, and T.-T. Wong, ``Deep halftoning with reversible binary
  pattern,'' in \emph{Proceedings of the IEEE/CVF International Conference on
  Computer Vision}, 2021, pp. 14\,000--14\,009.

\bibitem{TOG2021r3}
M.~Xie, M.~Xia, X.~Liu, C.~Li, and T.-T. Wong, ``Seamless manga inpainting with
  semantics awareness,'' \emph{ACM Transactions on Graphics (TOG)}, vol.~40,
  no.~4, pp. 1--11, 2021.

\bibitem{cvpr2021r4}
M.~Xie, M.~Xia, and T.-T. Wong, ``Exploiting aliasing for manga restoration,''
  in \emph{Proceedings of the IEEE/CVF Conference on Computer Vision and
  Pattern Recognition}, 2021, pp. 13\,405--13\,414.

\bibitem{tog2020r5}
M.~Xie, C.~Li, X.~Liu, and T.-T. Wong, ``Manga filling style conversion with
  screentone variational autoencoder,'' \emph{ACM Transactions on Graphics
  (TOG)}, vol.~39, no.~6, pp. 1--15, 2020.

\bibitem{tog2018r6}
L.~Zhang, C.~Li, T.-T. Wong, Y.~Ji, and C.~Liu, ``Two-stage sketch
  colorization,'' \emph{ACM Transactions on Graphics (TOG)}, vol.~37, no.~6,
  pp. 1--14, 2018.

\bibitem{r7han}
C.~Han, Q.~Wen, S.~He, Q.~Zhu, Y.~Tan, G.~Han, and T.-T. Wong, ``Deep
  unsupervised pixelization,'' \emph{ACM Transactions on Graphics (SIGGRAPH
  Asia 2018 issue)}, vol.~37, no.~6, pp. 243:1--243:11, November 2018.

\bibitem{r8liu2015closure}
X.~Liu, T.-T. Wong, and P.-A. Heng, ``Closure-aware sketch simplification,''
  \emph{ACM Transactions on Graphics (TOG)}, vol.~34, no.~6, pp. 1--10, 2015.

\bibitem{r9qu2008richness}
Y.~Qu, W.-M. Pang, T.-T. Wong, and P.-A. Heng, ``Richness-preserving manga
  screening,'' \emph{ACM Transactions on Graphics (TOG)}, vol.~27, no.~5, pp.
  1--8, 2008.

\bibitem{r10wang2006deringing}
G.~Wang, T.-T. Wong, and P.-A. Heng, ``Deringing cartoons by image analogies,''
  \emph{ACM Transactions on Graphics (TOG)}, vol.~25, no.~4, pp. 1360--1379,
  2006.

\bibitem{shesh2008efficient}
A.~Shesh and B.~Chen, ``Efficient and dynamic simplification of line
  drawings,'' in \emph{Computer Graphics Forum}, vol.~27, no.~2.\hskip 1em plus
  0.5em minus 0.4em\relax Wiley Online Library, 2008, pp. 537--545.

\bibitem{noris2012smart}
G.~Noris, D.~S{\`y}kora, A.~Shamir, S.~Coros, B.~Whited, M.~Simmons,
  A.~Hornung, M.~Gross, and R.~Sumner, ``Smart scribbles for sketch
  segmentation,'' in \emph{Computer Graphics Forum}, vol.~31, no.~8.\hskip 1em
  plus 0.5em minus 0.4em\relax Wiley Online Library, 2012, pp. 2516--2527.

\bibitem{modelbasedDRL}
A.~Plaat, W.~Kosters, and M.~Preuss, ``Model-based deep reinforcement learning
  for high-dimensional problems, a survey,'' \emph{arXiv preprint
  arXiv:2008.05598}, 2020.

\bibitem{ddpg}
T.~P. Lillicrap, J.~J. Hunt, A.~Pritzel, N.~Heess, T.~Erez, Y.~Tassa,
  D.~Silver, and D.~Wierstra, ``Continuous control with deep reinforcement
  learning,'' \emph{arXiv:1509.02971}, 2015.

\bibitem{konda2000actor}
V.~Konda and J.~Tsitsiklis, ``Actor-critic algorithms,'' \emph{Advances in
  neural information processing systems}, 2000.

\bibitem{l2p}
Z.~Huang, W.~Heng, and S.~Zhou, ``Learning to paint with model-based deep
  reinforcement learning,'' in \emph{Proceedings of the IEEE/CVF International
  Conference on Computer Vision (CVPR)}, 2019, pp. 8709--8718.

\bibitem{resnet}
K.~He, X.~Zhang, S.~Ren, and J.~Sun, ``Deep residual learning for image
  recognition,'' in \emph{Proceedings of the IEEE conference on computer vision
  and pattern recognition (CVPR)}, 2016, pp. 770--778.

\bibitem{BN}
S.~Ioffe and C.~Szegedy, ``Batch normalization: Accelerating deep network
  training by reducing internal covariate shift,'' in \emph{International
  conference on machine learning (ICML)}.\hskip 1em plus 0.5em minus
  0.4em\relax PMLR, 2015, pp. 448--456.

\bibitem{wn}
T.~Salimans and D.~P. Kingma, ``Weight normalization: A simple
  reparameterization to accelerate training of deep neural networks,''
  \emph{Advances in neural information processing systems (NIPS)}, vol.~29, pp.
  901--909, 2016.

\bibitem{adobe}
Adobe, ``Adobe illustrator,'' \emph{Image Trace, http://www.adobe.com/}, 2021.

\bibitem{potrace}
P.~Selinger, ``Potrace: a polygon-based tracing algorithm,'' \emph{Potrace
  (online), http://potrace. sourceforge. net/potrace. pdf (2009-07-01)},
  vol.~2, 2003.

\bibitem{TVCG2016manga}
C.-Y. Yao, S.-H. Hung, G.-W. Li, I.-Y. Chen, R.~Adhitya, and Y.-C. Lai, ``Manga
  vectorization and manipulation with procedural simple screentone,''
  \emph{IEEE transactions on visualization and computer graphics}, vol.~23,
  no.~2, pp. 1070--1084, 2016.

\bibitem{cairosvg}
Kozea, ``Cairosvg,'' \emph{https://cairosvg.org/}.

\bibitem{ssim}
Z.~Wang, A.~C. Bovik, H.~R. Sheikh, and E.~P. Simoncelli, ``Image quality
  assessment: from error visibility to structural similarity,'' \emph{IEEE
  Transactions on Image Processing (TIP)}, vol.~13, no.~4, pp. 600--612, 2004.

\bibitem{favreau2016fidelity}
J.-D. Favreau, F.~Lafarge, and A.~Bousseau, ``Fidelity vs. simplicity: a global
  approach to line drawing vectorization,'' \emph{ACM Transactions on Graphics
  (TOG)}, vol.~35, no.~4, pp. 1--10, 2016.

\bibitem{mckinley1998cubic}
S.~McKinley and M.~Levine, ``Cubic spline interpolation,'' \emph{College of the
  Redwoods}, vol.~45, no.~1, pp. 1049--1060, 1998.

\bibitem{tcsvtliu2015making}
Z.~Liu, H.~Li, W.~Zhou, T.~Rui, and Q.~Tian, ``Making residual vector
  distribution uniform for distinctive image representation,'' \emph{IEEE
  Transactions on Circuits and Systems for Video Technology (TCSVT)}, vol.~26,
  no.~2, pp. 375--384, 2015.

\bibitem{tcsvtyao2016resolution}
C.-Y. Yao, K.-Y. Chen, H.-N. Guo, C.-C. Li, and Y.-C. Lai, ``Resolution
  independent real-time vector-embedded mesh for animation,'' \emph{IEEE
  Transactions on Circuits and Systems for Video Technology (TCSVT)}, vol.~27,
  no.~9, pp. 1974--1986, 2016.

\end{thebibliography}

\begin{IEEEbiography}[{\includegraphics[width=1in]{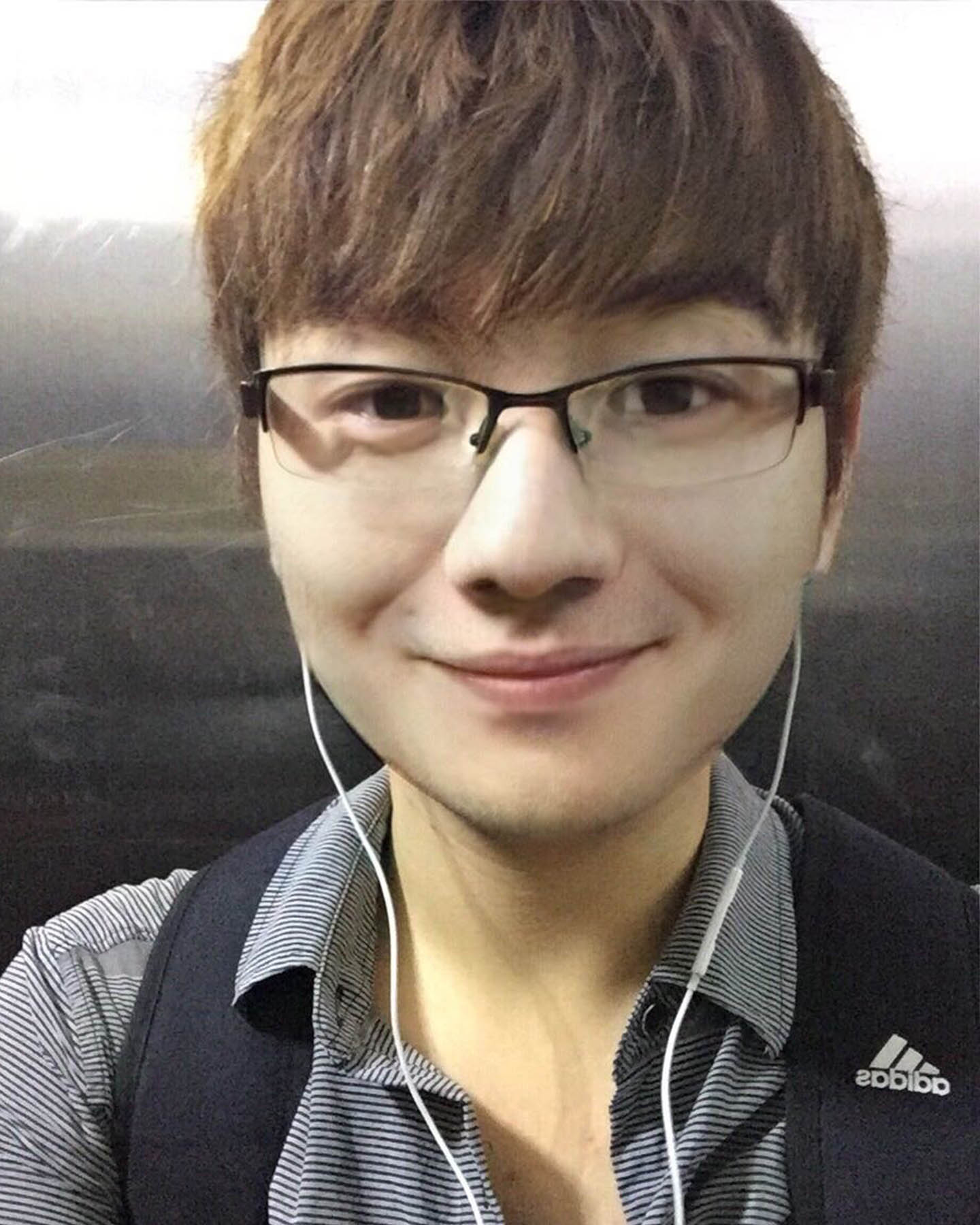}}]{Hao Su}
is a Ph.D. student in State Key Laboratory of Virtual Reality Technology and Systems, Beihang University, China. He received the B.E. degree in Computer Science and Technology from Zhengzhou University, in 2016. His research interests are in deep learning, computer graphics, and image processing.
\end{IEEEbiography}

\begin{IEEEbiography}[{\includegraphics[width=1in,height=1.25in,clip,keepaspectratio]{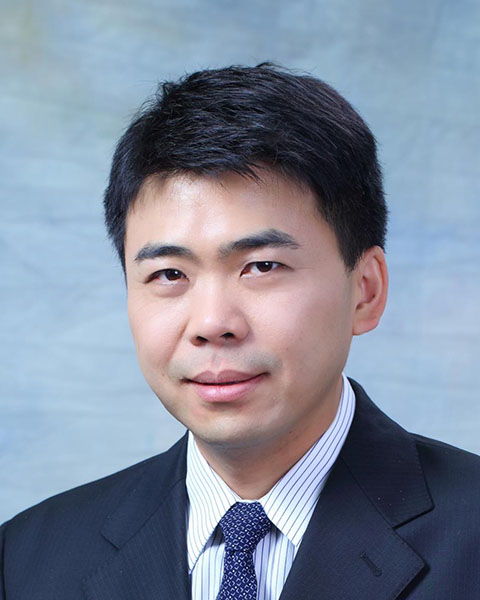}}]{Xuefeng Liu}
 received the M.S. and Ph.D. degrees from the Beijing Institute of Technology, China, and the University of Bristol, United Kingdom, in 2003 and 2008, respectively. He was an associate professor at the School of Electronics and Information Engineering in the HuaZhong University of Science and Technology, China from 2008 to 2018. He is currently an associate professor at the School of Computer Science and Engineering, Beihang University, China. His research interests include wireless sensor networks, distributed computing and in-network processing. He has served as a reviewer for several international journals/conference proceedings.
\end{IEEEbiography}

\begin{IEEEbiography}[{\includegraphics[width=1in,height=1.25in,clip,keepaspectratio]{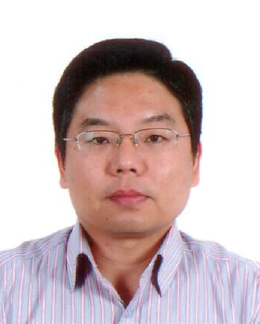}}]{Jianwei Niu}
 received the M.S. and Ph.D. degrees in computer science from Beihang University, Beijing, China, in 1998 and 2002, respectively. He was a visiting scholar at School of Computer Science, Carnegie Mellon University, USA from Jan. 2010 to Feb. 2011. He is a professor in the School of Computer Science and Engineering, BUAA, and an IEEE senior member. His current research interests include mobile and pervasive computing, mobile video analysis.
\end{IEEEbiography}

\begin{IEEEbiography}[{\includegraphics[width=1in,height=1.25in,clip,keepaspectratio]{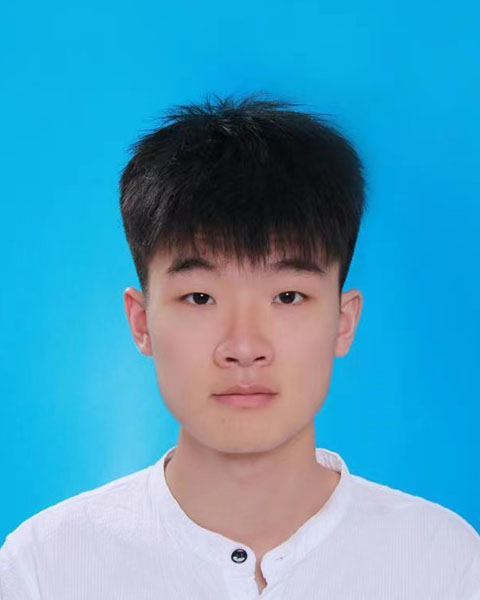}}]{Jiahe Cui}
is currently working toward the PhD degree in college of computer and science at Beihang University, Beijing, China. His research interests are multi-sensor SLAM and perceptual algorithms in autonomous driving.
\end{IEEEbiography}

\begin{IEEEbiography}[{\includegraphics[width=1in,height=1.25in,clip,keepaspectratio]{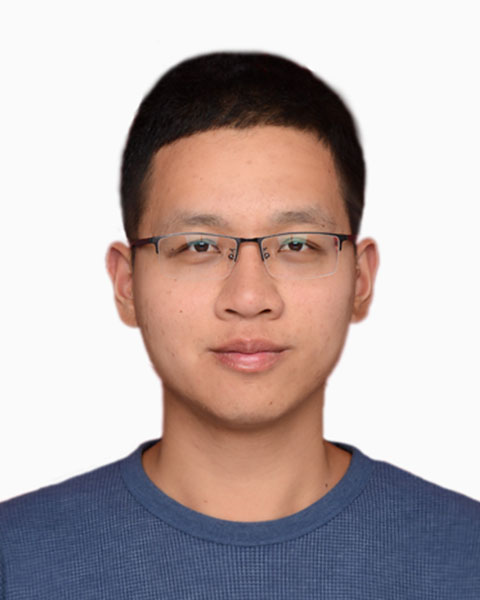}}]{Ji Wan}
is currently working toward the PhD degree at the State Key Laboratory of Software Development Environment, Beihang University. His research interests include distributed systems and blockchain.
\end{IEEEbiography}

\begin{IEEEbiography}[{\includegraphics[width=1in,height=1.25in,clip,keepaspectratio]{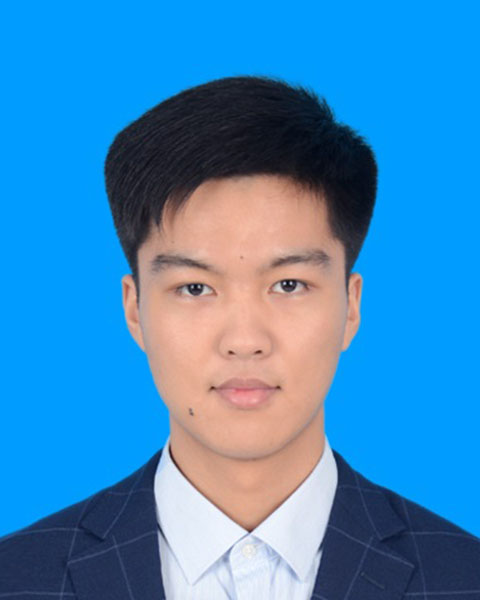}}]{Xinghao Wu}
is currently working toward the PhD degree in the school of computer science and engineering, Beihang University. His research interests are federated learning and distributed machine learning.
\end{IEEEbiography}
\end{document}